\documentclass[10pt,twocolumn,letterpaper]{article}

\usepackage[pagenumbers]{cvpr} 
\usepackage{float}

\usepackage{multirow}
\usepackage{multicol}
\usepackage{makecell}

\definecolor{cvprblue}{rgb}{0.21,0.49,0.74}
\usepackage[pagebackref,breaklinks,colorlinks,allcolors=cvprblue]{hyperref}

\newcommand{\yr}[1]{\textcolor{black}{#1}}
\newcommand{\szm}[1]{\textcolor{black}{#1}}
\newcommand{\yrn}[1]{\textcolor{black}{#1}}
\newcommand{\szmn}[1]{\textcolor{black}{#1}}
\newcommand{\szmnn}[1]{\textcolor{black}{#1}}
\newcommand{\tyz}[1]{\textcolor{black}{#1}}
\newcommand{\hut}[1]{\textcolor{black}{#1}}
\newcommand{\CR}[1]{\textcolor{black}{#1}}

\def\paperID{3030} 
\def\confName{CVPR}
\def\confYear{2025}

\begin{document}

\title{A$^\text{T}$A: Adaptive Transformation Agent for Text-Guided \\Subject-Position Variable Background Inpainting}

\author{
Yizhe Tang$^{1}$\footnotemark[1] ~~
Zhimin Sun$^{1,2}$\footnotemark[1] ~~
Yuzhen Du$^{1,2}$ ~~
Ran Yi$^{1}$\footnotemark[2] ~~
Guangben Lu$^{2}$\\
Teng Hu$^{1}$ ~~
Luying Li$^{2}$ ~~
Lizhuang Ma$^{1}$ ~~
Fangyuan Zou$^{2}$ \\ 
$^1$Shanghai Jiao Tong University \quad $^2$Tencent
}

\maketitle

{
  \renewcommand{\thefootnote}%
    {\fnsymbol{footnote}}
  \footnotetext[1]{Equal contribution. \quad $\{$tangyizhe, zhimin.sun$\}$@sjtu.edu.cn}
  \footnotetext[2]{Corresponding author. \quad ranyi@sjtu.edu.cn}
}

\begin{abstract}
Image inpainting aims to fill the missing region of an image.
Recently, there has been a surge of interest in foreground-conditioned background inpainting, a sub-task that \yr{fills} the background of an image while the foreground subject and associated text prompt are provided.
Existing background inpainting \yr{methods} typically 
\yr{strictly} preserve the subject's original position from the source image\yr{,} 
resulting in inconsistencies between the subject and the generated background.
To address this challenge, we propose \yr{a new task,} the ``Text-Guided Subject-Position Variable Background Inpainting'', which aims to dynamically adjust the subject position to \yr{achieve a harmonious relationship between the subject and} 
the inpainted background, and propose the Adaptive Transformation Agent (A$^\text{T}$A) for this task.
Firstly, we design a PosAgent Block that adaptively predicts an appropriate displacement based on given features to achieve variable subject-position.
Secondly, we design the Reverse Displacement Transform (RDT) module, which arranges multiple PosAgent blocks in a reverse structure, to transform hierarchical feature maps from deep to shallow based on semantic information.
Thirdly, we equip A$^\text{T}$A with a Position Switch Embedding to control whether the subject's position in the generated image is adaptively predicted or fixed.
Extensive comparative experiments validate the effectiveness of our A$^\text{T}$A approach, which not only demonstrates superior inpainting capabilities in subject-position variable inpainting, but also ensures good performance on subject-position fixed inpainting.
\end{abstract}
    
\section{Introduction}
\label{sec:intro}

\yr{Image inpainting~\cite{xiang2023inpainting} aims to fill the missing region in an image, which is an important research topic in computer vision.
With the development of \szm{text-to-image (T2I) diffusion models~\cite{rombach2021highresolution, peebles2023scalable, li2024hunyuan}, text-guided inpainting methods~\cite{avrahami2022blended, avrahami2023blended, xie2023smartbrush, ju2024brushnet, zhuang2023powerpaint,hu2024anomalydiffusion}} have achieved promising progress\yrn{, which fill a user-specified region under the inputs of an image, a masked region, and a text prompt, and achieve inpainting results consistent with the text prompt.}
In most inpainting scenarios, the \yrn{inpainted} regions are foreground regions, and some parts of the foreground human or objects are restored or edited according to the text guidance~\cite{avrahami2022blended, zhang2023adding, rombach2021highresolution, xie2023smartbrush, ju2024brushnet, zhuang2023powerpaint}.
However, in this paper, we focus on the inpainting scenarios where the foreground region is given and the \tyz{whole} background region is \yrn{to be inpainted}~\cite{pinco, zhang2024transparent, xie2024anywhere}, and aim to fill the background region according to the text prompt, which has wide applications in advertisement design, product promotional design, \textit{etc.}
}

\begin{figure}
    \centering
    \includegraphics[width=0.95\linewidth]{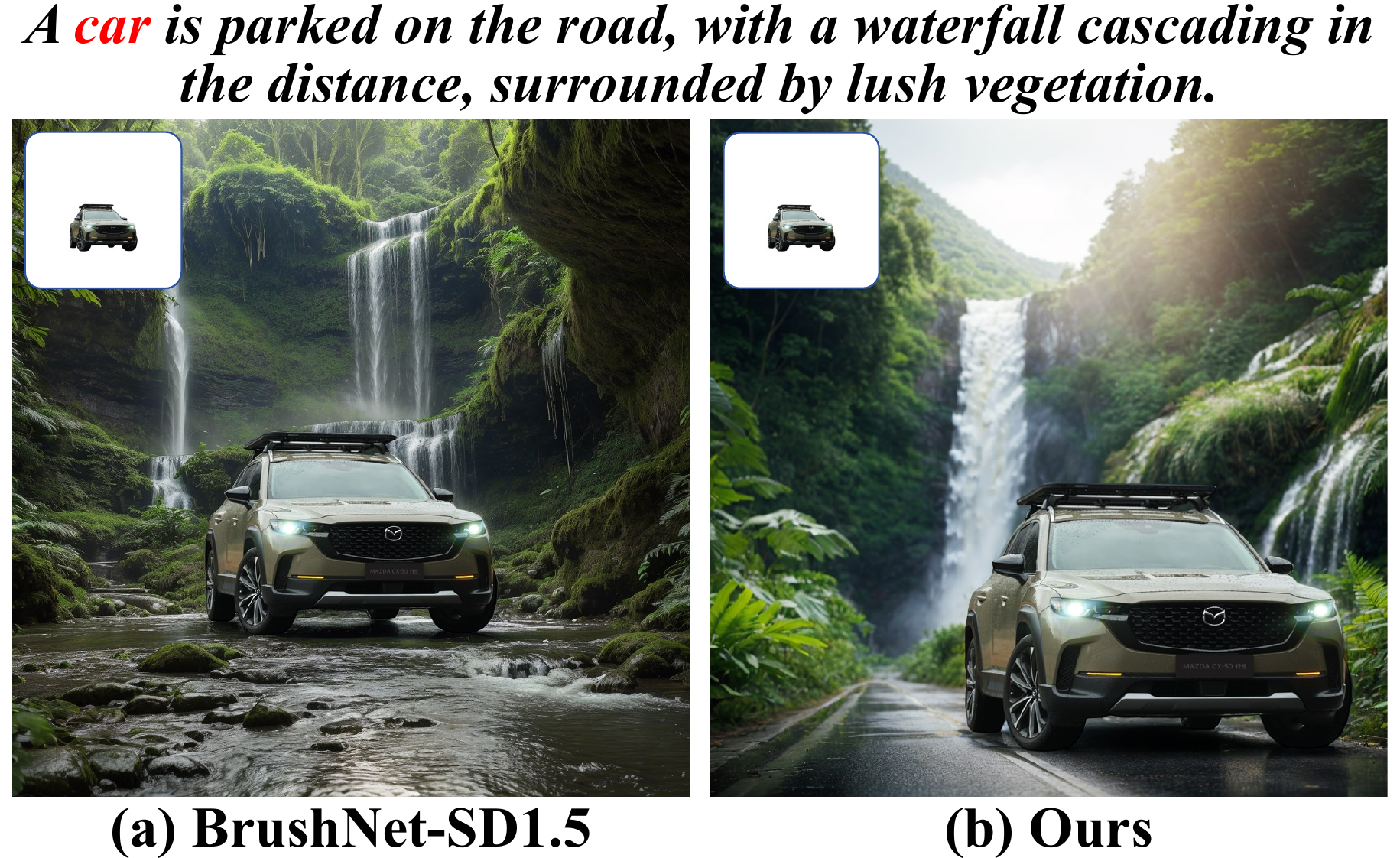}
    \vspace{-0.1cm}
    \caption{\yrn{For foreground-conditioned background inpainting, (a)} fixing the \hut{object position specified by}  the \yr{input} image \yr{(left-top)} may contradict the generated background; \yrn{(b) while our model achieves subject-position variable background inpainting, adaptively determines a suitable location for the subject, and generates an image with a harmonious subject-background relationship.} 
    }
    \label{fig:car-comp}
    \vspace{-0.6cm}
\end{figure}

\yr{When filling the \yrn{specified masked} regions, existing inpainting methods~\cite{xie2023smartbrush, zhuang2023powerpaint, ju2024brushnet} emphasize the \textit{preservation of the unmasked \yrn{(given)} region} as much as possible, requiring pixel-\yrn{level} strong consistency for the unmasked region.
\szm{However, in the case of \szmn{foreground-conditioned \yrn{background} inpainting~\cite{xie2024anywhere, pinco, zhang2024transparent}},} where \yrn{the foreground is given and the background is to be inpainted,} 
this unmasked region preservation constraint can sometimes be \szm{overpowering and result in problems.}
Given a subject image, a \yrn{background} mask, 
and a text prompt describing the desired background, 
\yrn{filling the content in the background while requiring the foreground strictly unchanged}
will limit the \yrn{subject's} position to the same position in the original image.}
However, 
\yr{in background inpainting scenarios in advertisement and art designs,}
\szm{the scale and position of the subject} \yr{in the generated image often}
cannot be determined before generation.
\yr{More importantly, directly fixing the position of the \yrn{subject} to that in the original image may contradict the generated background. 
\emph{E.g.}, given an image of a car in the center, and a text prompt ``\tyz{A car is parked on the road, with a waterfall cascading in the distance...}", if fixing the car position \yrn{to that in the original image}, the 
\yrn{inpainted image may not fully draw the desired background}
(\yrn{the car is on mud instead of on road in} Fig.~\ref{fig:car-comp}\yrn{(a)}).}

Taking these considerations into account, we propose a new task for this scenario, \szmn{the \textbf{Text-Guided Subject-Position Variable Background Inpainting} task}, \yr{where the position of the \szmn{subject} can \szm{be adaptively varied}}. 
The core of this task is the ability to \szm{adaptively determine a suitable location of subject,} based on the \szmn{subject} information \yr{in the original image} and \yr{the} text \yr{prompt describing the desired background}, \yr{thereby generating an} inpaint\yr{ed image} that 
\yr{\yrn{has a} harmonious positional relationship between the \szmn{subject} and the inpainted background}.

\szm{For the \szmn{Text-Guided Subject-Position Variable Background Inpainting} task, we propose the \szmn{\textbf{Adaptive Transformation Agent (A$^\text{T}$A)}}.}
\szmnn{\yr{Firstly, t}o \yr{achieve} variable subject-position, we \yr{design a} \textbf{PosAgent \yr{Block}} to adaptively predict an appropriate displacement \yr{based on given features, to move the subject to} the appropriate position.
\yr{Specifically, the spatial transformation is applied at the feature level, where the PosAgent predicts a pair of displacement transformation parameters, which are then used to transform multi-scale subject features by spatial feature transform (SFT).}
\yr{Secondly, to transform hierarchical feature maps from deep to shallow based on semantic information, we design the \textbf{Reverse Displacement Transform (RDT) module}, which arranges multiple PosAgent blocks in a reverse structure, \emph{i.e.,} the output of the first PosAgent block transforms the deepest feature map, while the last PosAgent block transforms the shallowest feature map. This reverse transform structure effectively alleviates the subject deformation problems of separate or sequential structures, while achieving subject-background harmonious position prediction.}
\yr{Thirdly, to allow the model to both adaptively move the subject and maintain the subject's position,}
we \yr{equip the A$^\text{T}$A
with a} \textbf{Position Switch Embedding},
\yr{which controls}
whether the position of the subject in the generated image is adaptively \yr{predicted} or fixed.
\yr{By setting the position switch embedding, users can flexibly switch between subject-position variable and subject-position fixed inpainting.}
\yr{To train A$^\text{T}$A with the position switch embedding,} we use a hybrid training \yr{strategy}, where half of the data uses \yr{variable} position samples and half uses \yr{fixed} position ones.}

\szmn{Extensive comparative experiments validate the effectiveness of our A$^\text{T}$A approach, which not only demonstrates superior inpainting capabilities \yr{in} subject-position variable \yr{inpainting,} 
but also ensures \yr{good} performance on \yr{subject-position} fixed inpainting.}

We summarize our contributions as follows:
\begin{enumerate}
    \item We propose \szm{a novel \szmn{Text-Guided Subject-Position Variable Background Inpainting} task, which \yr{aims at} generat\yr{ing} inpaint\yr{ed} images with a harmonious positional relationship between the \szmn{subject} and the inpainted background.}
    \item We propose \szmn{Adaptive Transformation Agent \yr{(A$^\text{T}$A)}}, \szmn{which adaptively shifts and scales the subject \yr{to an appropriate position} through the Reverse Displacement Transform \yr{(RDT)} module}. \yr{RDT consists of multiple PosAgent blocks that predict the displacement transformation parameters, which are then applied to transform hierarchical features by SFT, and arranges the PosAgent blocks in a novel reverse structure, effectively alleviating subject deformation.}
    \item We 
    \yr{equip the A$^\text{T}$A with a Position Switch Embedding, which controls whether the position of the subject is adaptively predicted or fixed, making the model both can adaptively move the subject and maintain the subject's position. An end-to-end hybrid training strategy is designed to train A$^\text{T}$A with the position switch embedding.}
\end{enumerate}

\section{Related Work}
\label{sec:related}

\subsection{Image Inpainting}
With the development of T2I diffusion models~\cite{rombach2021highresolution}, great progress has been made in image inpainting~\cite{xiang2023inpainting, rombach2021highresolution, avrahami2022blended, avrahami2023blended, xie2023smartbrush, ju2024brushnet, zhuang2023powerpaint,hu2024anomalydiffusion, xu2023hierarchical, du2024ld}.
Blended Latent Diffusion~\cite{avrahami2022blended, avrahami2023blended} uses separate denoising for masked and unmasked areas to achieve inpainting, and SD-Inpainting~\cite{rombach2021highresolution} and ControlNet-Inpainting~\cite{zhang2023adding} further refine by fine-tuning StableDiffusion~\cite{rombach2021highresolution} models.
SmartBrush~\cite{xie2023smartbrush} enhances model understanding of complex scenes by training with pairs of descriptive objects and corresponding masks.
PowerPaint~\cite{zhuang2023powerpaint} combines image inpainting with image removal for better text-image alignment.
BrushNet~\cite{ju2024brushnet} employs a versatile two-branch network architecture to inject features from masked images and ensure consistent inpainting results.

Existing inpainting methods, though effective, sometimes fail to integrate backgrounds seamlessly, due to difficulties in matching the fixed subject with the varying background context during generation.
In comparison, our A$^\text{T}$A framework overcomes this by using the Reverse Displacement Transform module to dynamically adjust the position and scale of the subject, ensuring a natural integration of subject and background in the inpainted images.

\subsection{Controllable Image Generation}

As the 
image generation
model continues to evolve, there is a growing emphasis on controllable image generation~\cite{FeditNet++, sun2023contrastive, sun2024rethinking} within the research community.
Works that involve adapter networks~\cite{zhang2023adding, ye2023ipa, mou2024t2i} control content by integrating side-branch adapter networks, which can be used for inpainting tasks after fine-tuning.
However, these works can hardly achieve the appropriate subject position, requiring manual attempts at different positions and zoom levels.

Personalized Image Generation~\cite{ruiz2023dreambooth, shi2024instantbooth, li2024photomaker, li2024blip, wei2023elite} generates new images by preserving the identity information from the original image, yielding artistic and stylized images with a high degree of creative freedom.
Nevertheless, these personalized models generate images based solely on textual content, lacking the capability for controlled generation in conjunction with a subject image.

Layout-to-Image, a multi-stage task for controlled image generation using layouts, also meets challenges when applied to our task. In the first stage, users must specify the positions of objects and backgrounds to generate a layout~\cite{feng2024layoutgpt, zheng2023layoutdiffusion, avrahami2023spatext}, and controllable T2I methods~\cite{li2023gligen, xie2023boxdiff, chefer2023attendandexcite} are used in the second stage to generate the image.
Although these methods have achieved satisfactory results, they can not realize pixel-level control of the subject, which is crucial for our task. Additionally, the layout generation process is complicated, resulting in unreliable generation of inpainting images, further highlighting the need for our A$^\text{T}$A.

\section{Preliminaries}
\label{sec:pre}

\subsection{Diffusion Models}

Diffusion models~\cite{rombach2021highresolution} encompass forward and backward processes. In the forward diffusion process, a clean sample $ x_0 $ is converted into a noise sample $ x_t $ by adding Gaussian noise $\epsilon$, which can be expressed as follows:
\begin{equation}
    x_t = \sqrt{\alpha_t} x_0 + \sqrt{1 - \alpha_t} \epsilon, \ \epsilon \sim \mathcal{N}(0,1),
\end{equation}
where $x_t$ represents the noised feature at step $ t $ and $ \alpha_t $ denotes a hyper-parameter of the noise level.
In the backward denoising process, starting with the noise $ x_T $, a trainable network $ \epsilon_{\theta} $ predicts the noise at each step $ t $ leveraging conditional $ c $, and derives original data $ x_0 $ after T iterations. 

The diffusion model training process aims to optimize the denoising network $\epsilon_{\theta}$ so that it can accurately predict the noise given the condition $ c $. The training loss function is formulated as:
\begin{equation}
    \mathcal{L} =\mathbb{E}_{x_0, c, \epsilon\sim\mathcal{N}(0,1), t} \left[ \| \epsilon - \epsilon_{\theta}(x_t, t, c) \|^2_2 \right].
\end{equation}
This loss function quantifies the discrepancy between the predicted noise and the actual noise, thereby guiding the optimization process.

\subsection{Hunyuan-DiT}
\label{sec:dit}

Hunyuan-DiT~\cite{li2024hunyuan} is the base model of Adaptive Transformation Agent, which is an open-source enhanced iteration of DiT~\cite{peebles2023scalable}, a transformer-based diffusion model operating on latent patches.
Hunyuan-DiT integrates the text condition with the diffusion model through cross-attention~\cite{rombach2021highresolution}, extends Rotary Positional Embedding (RoPE)~\cite{su2024rope} to a $2$D format, which encodes absolute and relative positional dependencies, and expands its capabilities to the image domain.
For text encoding, Hunyuan-DiT merges mT5~\cite{xue2020mt5} and CLIP~\cite{radford2021clip}, leveraging their complementary strengths to enhance the precision and creativity of the T2I generation.

\section{Text-Guided Subject-Position Variable Background Inpainting}
\label{sec:method}

\begin{figure*}[t]
    \centering
    \includegraphics[width=0.93\linewidth]{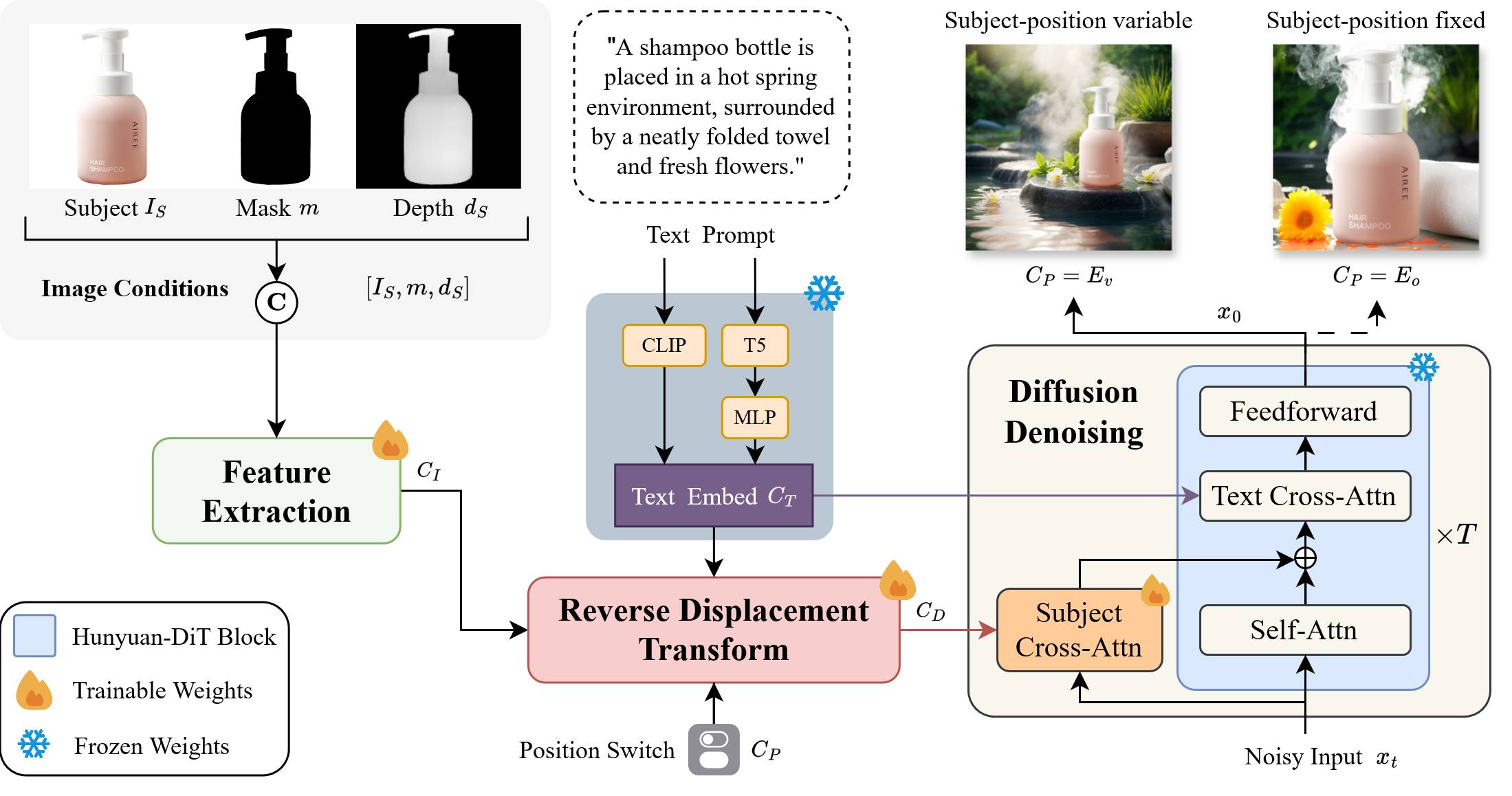}
    \vspace{-0.3cm}
    \caption{\textbf{Adaptive Transformation Agent (A$^\text{T}$A)} comprises $4$ modules: Feature extraction, \textbf{Reverse displacement transform}, Feature fusion, and Diffusion denoising. We use Hunyuan-DiT~\cite{li2024hunyuan} as the base model, and mainly develop the subject feature extraction, displacement transformation prediction, and displaced feature injection mechanisms to achieve \textit{subject-position variable inpainting}. We also \yrn{design a} \textbf{position switch embedding} to control whether the position of the subject in the generated image is adaptively predicted or fixed.}
    \label{fig:ata}
    \vspace{-0.5cm}
\end{figure*}

\subsection{\yr{Task} Definition}
A variety of tasks characterize the domain of image inpainting,
which can be broadly categorized into Text-guided object inpainting, Shape-guided object inpainting, Context-aware image inpainting, and Outpainting, as detailed in previous studies~\cite{xiang2023inpainting, zhuang2023powerpaint}.
Recently, there has been a surge of interest in \emph{foreground-conditioned \yrn{background} inpainting}, a sub-task that involves filling in the background of an image while the foreground subject and associated text prompt are provided~\cite{pinco, zhang2024transparent, xie2024anywhere}. This \yr{foreground-conditioned inpainting task} typically fixes the position of the subject to match that in the original \yr{subject} image. 
However, in applications such as advertising and artistic design, the position and scale of the subject within the generated image are often undetermined \yr{before} the generation process. Moreover, \yr{strictly} adhering to the original image's subject position can \yr{conflict} with the newly generated background, resulting in 
\yr{inconsistencies between the subject and background.} 

To \yr{address} this issue, we introduce a novel task: \textbf{Text-Guided Subject-Position Variable Background Inpainting}. Unlike \yr{the conventional} foreground-conditioned \yrn{background} inpainting \yr{task}~\cite{pinco, zhang2024transparent, xie2024anywhere}, this \yr{subject-position variable} task demands 
intelligently adjust\yr{ing} \yr{the} position and scale \yr{of the foreground subject} to achieve a harmonious relationship with the generated background, while maintaining the original shape and details of the subject.

\subsection{Adaptive Transformation Agent}
To address the challenges in the Text-Guided Subject-Position Variable Background Inpainting task, we propose a novel framework \textbf{Adaptive Transformation Agent} (A$^\text{T}$A), as shown in Fig.~\ref{fig:ata}, comprising $4$ modules: Feature extraction, Reverse displacement transform, Feature fusion, and Diffusion denoising.
\yr{We use Hunyuan-DiT~\cite{li2024hunyuan} as the base model,} and \szm{mainly develop the mechanisms of subject feature extraction, displacement prediction, and feature injection to achieve subject-position variable inpainting.}

\textbf{Feature extraction.} The \yr{inputs to our} A$^\text{T}$A framework \yr{consists of:} \yr{a subject} image $I$, 
\yr{a} background mask $m$, \yr{and a} text prompt $T$ \yr{(describing the desired background)}.
A$^\text{T}$A will preprocess the subject image \yr{$I$} for feature extraction. Specifically, we obtain the masked subject image \yr{by:}
\begin{equation}
    I_S = I \odot (1 - m).
    \label{eq:mask}
\end{equation}
\yr{Then, we} consider the mask $m$ and depth map $d_S$ \yr{(extracted by}~\cite{bhat2023zoedepth}) of the subject image as the 
\yr{condition maps,}
and use Swin-Transformer~\cite{liu2021swin} as an image encoder $\Phi$ to extract \yr{multi-scale subject} image feature $C_I$:
\begin{equation}\label{eq:swin}
    C_I = \Phi([I_S, m, d_S]),
\end{equation}
which can effectively capture the global and local features of the \yr{subject} through its hierarchical structure and sliding window mechanism.
Moreover, since our base model is a transformer-based structure \yr{(a text-to-image DiT)}, extract\yr{ing subject} features using Swin-Transformer can \yr{better} align \yr{with the base model} and \yr{allow for better fusion of the subject features through} 
attention operations~\cite{pinco}.
Meanwhile, \yr{the text features $C_T$ are} 
extracted by the CLIP~\cite{radford2021clip} and mT5~\cite{xue2020mt5} \yr{from the input text prompt $T$}.

\textbf{Reverse displacement transform.} Next, A$^\text{T}$A adopts our Reverse Displacement Transform (RDT) module, which \yr{consists of multiple PosAgent blocks that predict displacement transformation parameters and arrange them in a reverse structure} (\yr{detailed in} Sec.~\ref{sec:rdt}). The RDT module can effectively generate the multi-level displaced features $C_D$:
\begin{equation}\label{eq:rdt}
\setlength\abovedisplayskip{4pt}
\setlength\belowdisplayskip{4pt}
    C_D = \text{RDT} \left(C_I; C_T, C_P, t\right),
\end{equation}
where $C_P$ refers to the position switch embedding, \yr{which controls} 
whether 
the position of the subject \yr{in the generated image is adaptively predicted or fixed (detailed in Sec.~\ref{sec:position_switch});} 
\yr{and $t$ represents the \yrn{time embedding} in diffusion model.}

\textbf{Feature fusion \yrn{by subject cross-attention}.} Afterward, A$^\text{T}$A injects \yr{the displaced} feature\yr{s $C_D$} into 
\yr{the base T2I DiT model}
and strengthens the model's adaptation to the subject's location through \yr{decoupled cross-attention mechanism}~\cite{ye2023ipa}. \yr{Specifically, an} additional \textbf{\yr{subject cross-}attention} \yr{layer is designed to insert the subject features, which calculates relationships between displaced subject features and latent. Then, different from conventional IP-Adapter, we follow the Self-IPA from~\cite{pinco} and integrate the subject cross-attention with the self-attention layer by}:
\begin{equation}
\setlength\abovedisplayskip{6pt}
\setlength\belowdisplayskip{6pt}
\begin{aligned}
    x_t = \alpha & \odot \text{Self-Attention}(x_t W\yr{_q}, x_t W\yr{_k}, x_t W\yr{_v}) \\
     + \beta & \odot \text{Cross-Attention}(x_t W\yr{_q'}, C_D W\yr{_k'}, C_D W\yr{_v'}),
\end{aligned}
\end{equation}
where $\alpha, \beta$ are learnable weights for attention blocks, and $W\yr{_q}, W\yr{_k}, W\yr{_v}$ are trainable parameters. 
\szm{This decoupled attention strategy} maintains the stability of the original model through the injection of adapters.
\yr{After} this \yr{module}, text features \yr{$C_T$} are \yr{injected} by the \yr{text} cross-attention module, which is consistent with \yr{the} base model~\cite{li2024hunyuan}:
\begin{equation}
\setlength\abovedisplayskip{4pt}
\setlength\belowdisplayskip{4pt}
    x_t = \text{Cross-Attention}(x_t W\yr{_q}, C_T W\yr{_k}, C_T W\yr{_v}).
\end{equation}

\textbf{Diffusion denoising.} After all these features are fused, A$^\text{T}$A performs a T-step denoising process as in \yr{conventional} diffusion models \szm{and obtains the inpainted image.}

\begin{figure}
    \centering
    \begin{subfigure}{\linewidth}
    \includegraphics[width=0.84\linewidth]{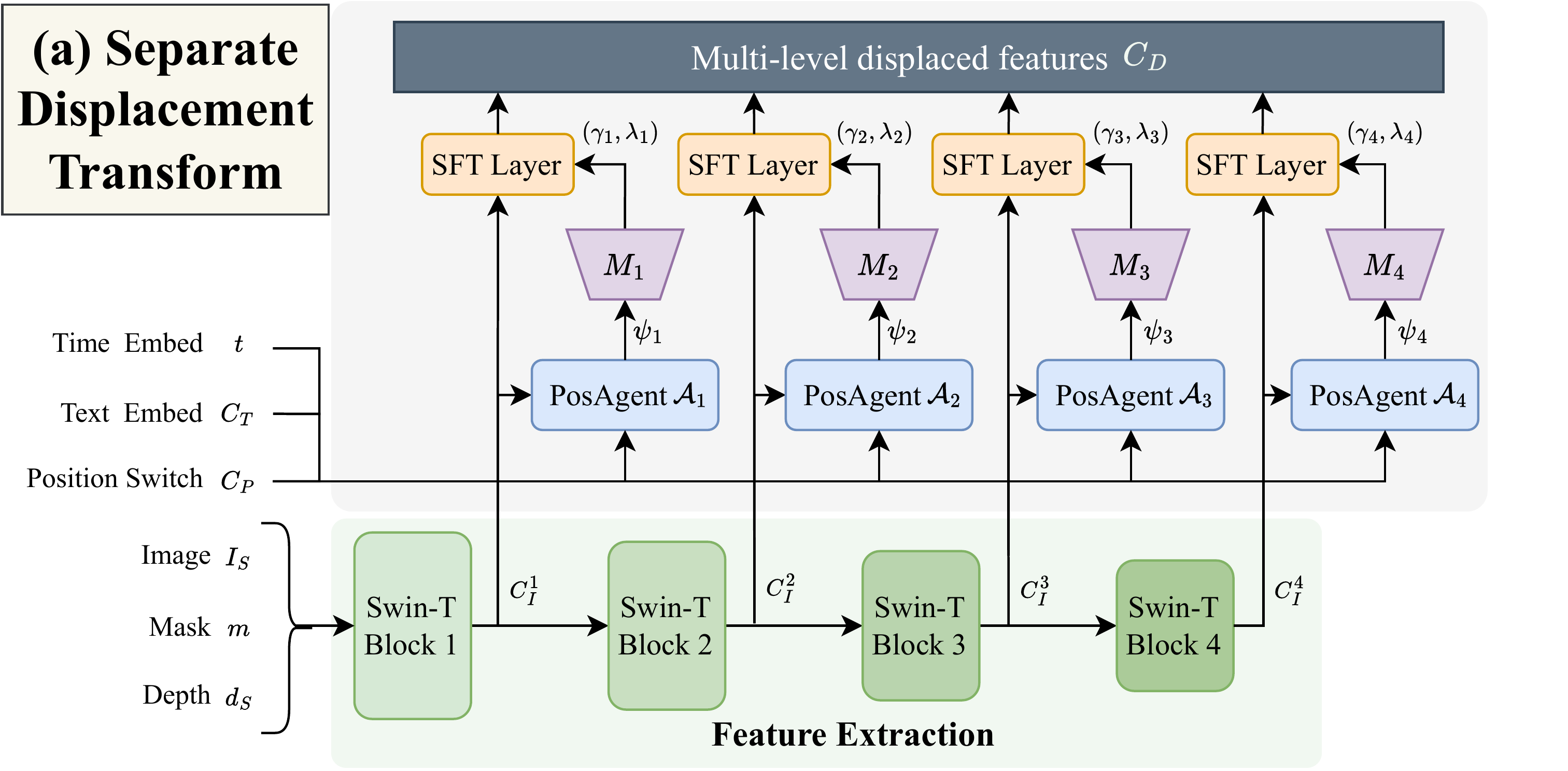}
    \end{subfigure}\\
    \begin{subfigure}{\linewidth}
    \includegraphics[width=\linewidth]{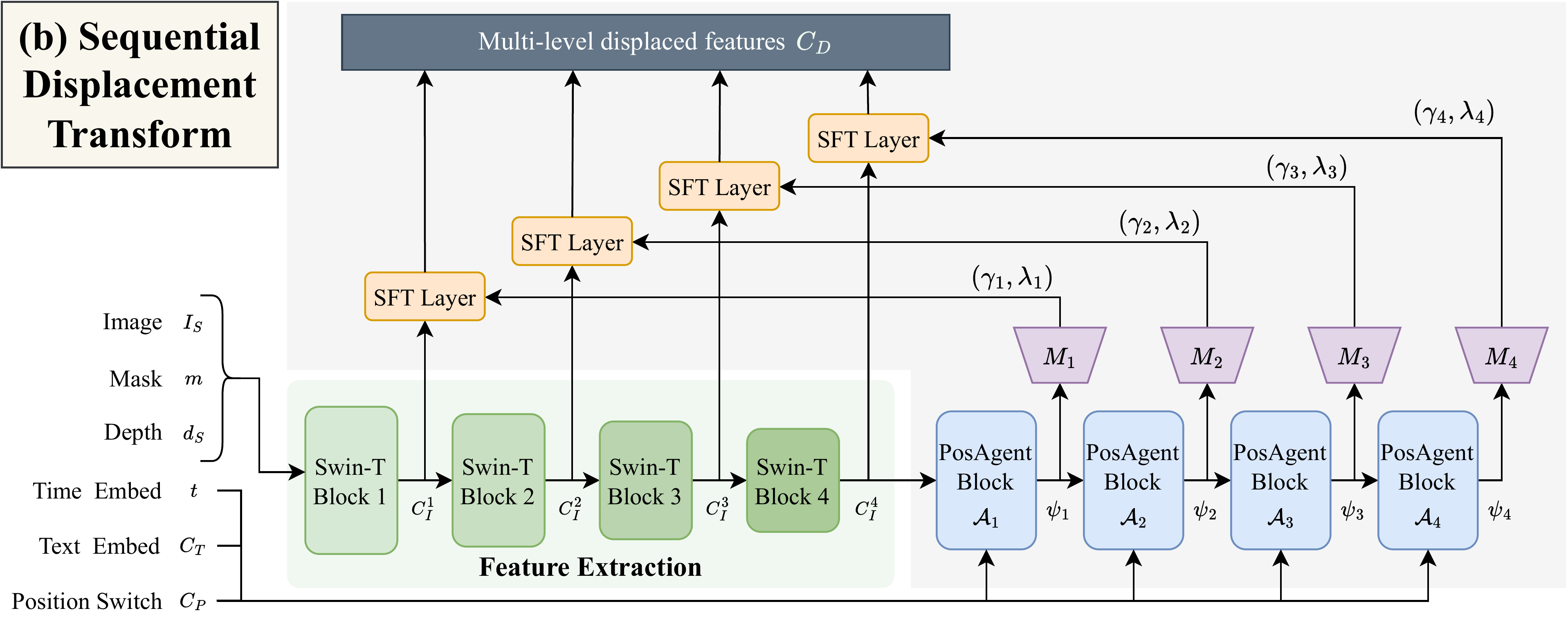}
    \end{subfigure}\\
    \begin{subfigure}{\linewidth}
    \includegraphics[width=\linewidth]{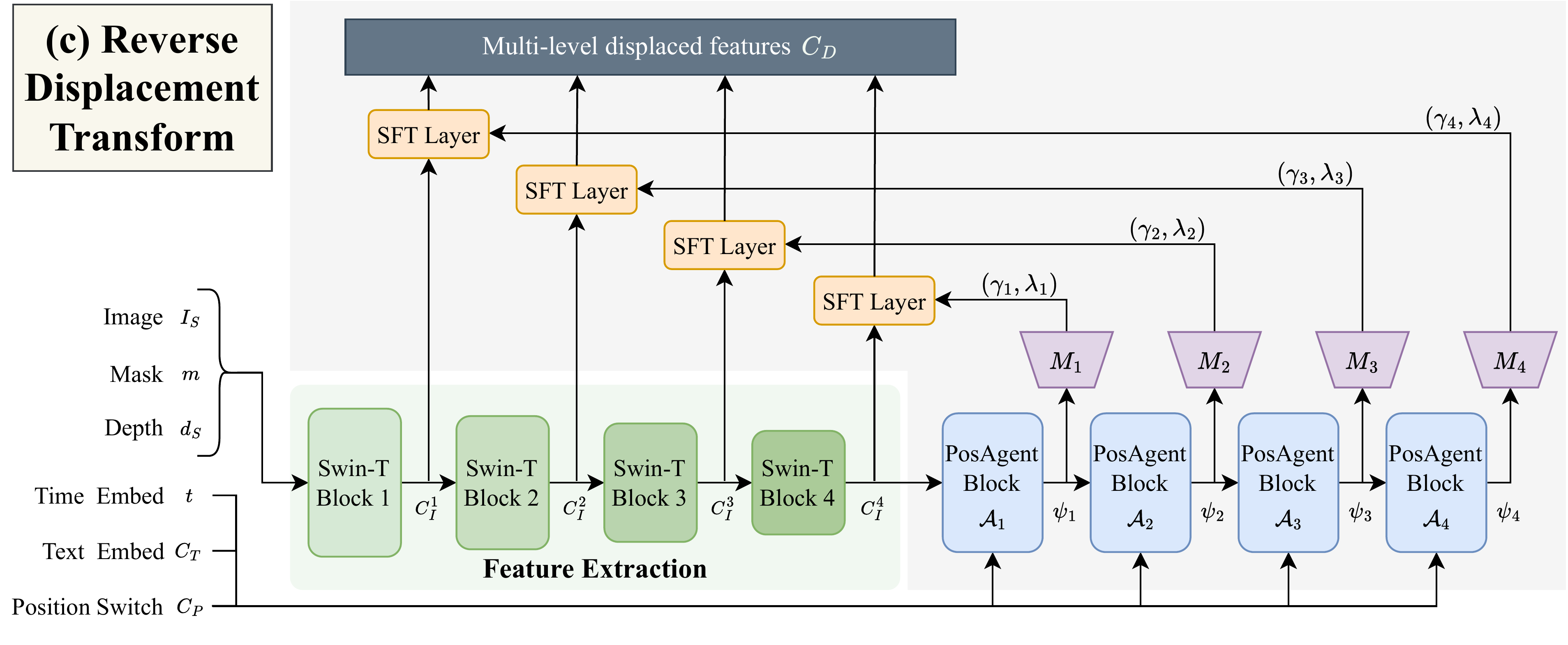}
    \end{subfigure}\\
    \vspace{-0.3cm}
    \caption{Comparison of different \yr{structures for} Displacement Transform. To transform the hierarchical feature maps from deep to shallow based on semantic information, we propose a novel Reverse Displacement Transform module.}
    \label{fig:rdt}
    \vspace{-0.5cm}
\end{figure}

\subsection{Reverse Displacement Transform}\label{sec:rdt}

Existing works~\cite{pinco, ju2024brushnet, zhuang2023powerpaint} use \szm{VAE}~\cite{kingma2013vae}
to extract condition\yr{s} from unmasked images for injection into the inpainting model.
Although effective for \yr{subject-position} fixed inpainting tasks, \szm{VAE is impractical} for our \yr{subject-position} variable inpainting task, \szm{which cannot} simultaneously extract sufficient image information and guarantee position independence.
\yrn{Therefore, apart from the subject feature extractor, a \textbf{PosAgent Block} \szm{is designed} to predict a suitable displacement for the subject to move it to a suitable position, and perform \szm{SFT} at the feature level,
\szm{and multiple consecutive PosAgent blocks are arranged
to} cope with multi-level features.
Different from} straightforward \yrn{ways} to \yrn{inject} position offset
\yrn{by} the Separate or Sequential Displacement Transform
(Fig.~\ref{fig:rdt}a,~\ref{fig:rdt}b),
\yrn{which} \yr{suffer from} subject deformation,
\yrn{our} \textbf{Reverse Displacement Transform} module
\szm{(Fig.~\ref{fig:rdt}c)}
transform\yr{s} hierarchical feature maps from deep to shallow based on the semantic information\yrn{, effectively alleviating subject deformation and achieving subject-background harmonious position prediction.}

\textbf{PosAgent blocks.}
\szm{
Firstly, to achieve variable subject position, we design a PosAgent \yr{block} $\mathcal{A}$ to adaptively predict a suitable displacement based on the given features.
\yr{The inputs of $\mathcal{A}$ consist of subject image feature $C_I$ (Eq.~\ref{eq:swin}), text feature $C_T$, position switch embedding $C_P$, and \yrn{time embedding} $t$, where $C_I$
contain multi-scale feature maps $\{C^1_I, \cdots, C^N_I\}$.
Given these inputs, $\mathcal{A}$ predicts a displacement transformation feature $\psi$, which will be mapped into transformation parameters.}
To cope with features at different \szm{scales,} we \yr{design} multiple consecutive PosAgent blocks $\yr{\{\mathcal{A}_1, \cdots, \mathcal{A}_N\}}$ to predict \yr{multiple} displacement transformation \yr{features}
$\yr{\{\psi_1, \cdots, \psi_N\}}$.
\yr{Specifically,} the \szm{deepest} feature map $C^N_I$ serves as the input to $\mathcal{A}_1$, capturing the most refined texture information~\cite{liu2021swin}\yr{, while the output displacement feature $\psi_{i-1}$ from $\mathcal{A}_{i-1}$} \szm{serves as input to $\mathcal{A}_{i}$:}}
\begin{equation}\label{eq:posagent}
    \psi_i = \left\{ 
    \begin{array}{ll}
        \mathcal{A}_i(C_I^N; C_T, C_P, t), & \text{if } i = 1, \\
        \mathcal{A}_i(\psi_{i-1}; C_T, C_P, t), & \text{otherwise.}
    \end{array}
    \right.
\end{equation}
For the output \yr{displacement feature} $\psi_i$ of each block $\mathcal{A}_i$, we \yr{then} apply a learnable mapping function $M_i$ that generates a pair of displacement transformation parameters, denoted as $\left( \gamma_i, \lambda_i \right)$, \yr{where $\gamma_i$ represents scale and $\lambda_i$ represents shift:}
\begin{equation}
    \left(\gamma_i, \lambda_i \right) = M_i(\psi_i).
\end{equation}
To perform better spatial transformation of hierarchical feature maps, we introduce a Spatial Feature Transform (SFT) layer~\cite{wang2018sft}.
\yr{After obtaining} 
\yr{the displacement transformation parameters $\{\left( \gamma_i, \lambda_i \right)\}_{i=1}^N$ from}
the \yr{multiple} PosAgent blocks, 
\yr{SFT layer transforms the hierarchical feature maps $\{C^1_I, \cdots, C^N_I\}$ through scaling and shifting.}

\textbf{Reverse\yr{-structure} transformation.}
To transform the hierarchical feature maps from deep to shallow based on the semantic information, we propose a novel reverse transformation \yr{module, which arranges multiple PosAgent blocks in a \textbf{reverse structure}.}
\szm{\emph{I.e.}, $\left(\gamma_1, \lambda_1 \right)$ \yr{are used to} transform the deepest feature map $C^N_I$, while $\left(\gamma_N, \lambda_N \right)$ will transform the shallowest feature map $C^1_I$ (shown in Fig.~\ref{fig:rdt}c). The process of using the $i$-th PosAgent's transformation parameters to transform the $(N\!+1\!-i)$-th feature map is formulated as:}
\begin{equation}
    \text{SFT} \left(C_I^{N\!+1\!-i}; \gamma_i, \lambda_i\right) = \gamma_{i} \odot C_I^{N\!+1\!-i} + \lambda_{i}.
\end{equation}
The reverse structure ensures that 
\yr{the deepest features that capture the most semantically rich information are processed first,}
and \yr{to compensate for the detailed information,} by \yr{gradually} combining deeper features with shallower ones \yr{that capture finer-grained details},
the model \yr{can} better understand the entire \yr{scene}.

Finally, we obtain the adaptive\yr{ly} \yr{displaced} multi-level features $C_D$ \yr{from} the RDT module, which will be subsequently injected into the base model:
\begin{equation}
    \text{RDT} \left(C_I ; C_T, C_P, t\right) = \left\{ \gamma_{i} \odot C_I^{N\!+1\!-i} + \lambda_{i}, i \in [1, N] \right\}.
\end{equation}

\subsection{Hybrid Training with Position Switch}
\label{sec:position_switch}

To provide users with more choices, we require A$^\text{T}$A to implement both inpainting tasks in the form of adaptive \yr{subject position} and \yr{fixed} subject position. 
\yr{To achieve this, we equip our A$^\text{T}$A with a \textbf{Position Switch Embedding}, enabling flexible switch between subject-position variable and fixed inpainting.}
\yr{To train the A$^\text{T}$A with the position switch embedding, we use an end-to-end \textbf{Hybrid Training Strategy}, where the}
training dataset includes samples \yr{from two} tasks, 
\yr{mixed together} in a $1\!:\!1$ ratio.

\begin{figure}
    \centering
    \begin{subfigure}{\linewidth}
    \centering
    \includegraphics[width=0.8\linewidth]{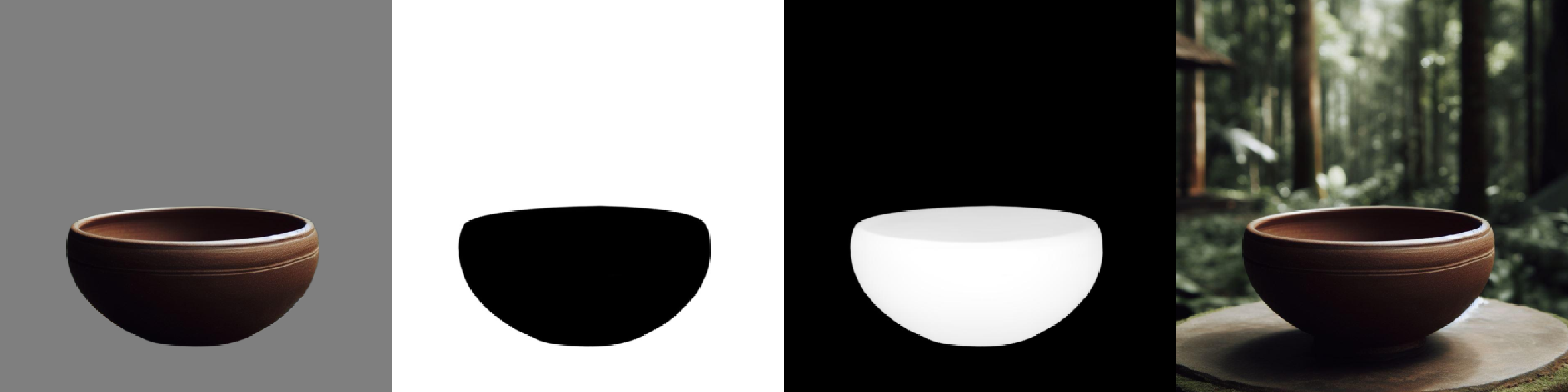}
    \caption{\yr{Subject-position} Fixed 
    Sample $I \in \mathcal{D}_o$.}
    \label{fig:origin_sample}
    \end{subfigure}
    \begin{subfigure}{\linewidth}
    \centering
    \includegraphics[width=0.8\linewidth]{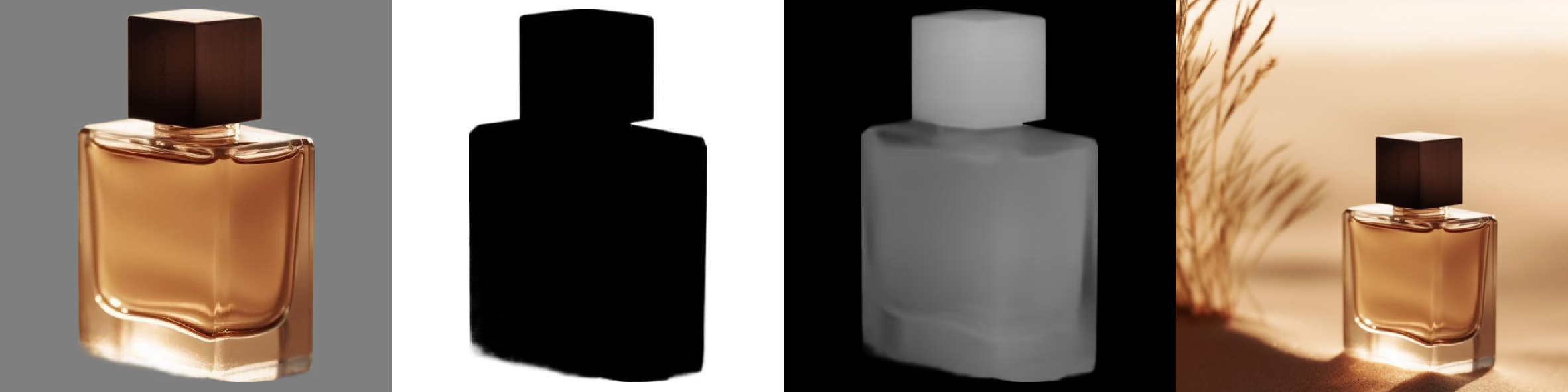}
    \caption{\yr{Subject-position} Variable 
    Sample $I \in \mathcal{D}_v$.}
    \label{fig:adaptive_sample}
    \end{subfigure}
    \vspace{-0.4cm}
    \caption{Training samples and corresponding image conditions. From left to right: masked subject image $I_S$, mask $m$, depth map $d_S$, ground truth image $I$.}
    \label{fig:sample}
    \vspace{-0.6cm}
\end{figure}

\textbf{Position switch.} \yr{To allow the model to both adaptively move the subject and maintain the subject's position}, we \yr{equip our A$^\text{T}$A framework with a} position switch \yr{embedding} $C_P$, \yr{which serves as a} 
``switch'' \yr{to control} whether 
\yr{the position of the subject is adaptively predicted or fixed.}
$C_P$ has $2$ states, $E_v$ for \yr{subject-position} variable \yr{inpainting,} 
while $E_o$ for \yr{subject-position} fixed \yr{inpainting,} 
where \yrn{both} $E_o$ and $E_v$ \yrn{are} \yr{learnable.} 
Along with text embedding $C_T$ and image embedding $C_I$, $C_P$ is injected into the condition side of the RDT (Eq.~\ref{eq:rdt})\yr{, serving as an input to the PosAgent when predicting the transformation parameters}.

\textbf{Hybrid training.}
\yr{We use an end-to-end hybrid training strategy to train the A$^\text{T}$A with the position switch embedding, where half of the training data uses variable position samples, and half uses fixed position samples.}
Besides the original training samples $\mathcal{D}_o$ (Fig.~\ref{fig:origin_sample}) for the fixed-position inpainting, we \yr{adopt} training samples $\mathcal{D}_v$ for our variable-position inpainting (Fig.~\ref{fig:adaptive_sample}).
\yrn{To construct $\mathcal{D}_v$, we process the subject image to center and enlarge the subject, use the centered image as the input, and use the original image (subject in a suitable position, away from the center) as the ground truth, to make the model adaptively learn a suitable subject position.}
\szm{$\mathcal{D}_v$ and $\mathcal{D}_o$ are mixed in a $1\!:\!1$ ratio.}
For \szm{variable-position} inpainting, we set the position switch embedding to $E_v$ and use $I \in \mathcal{D}_v$ as the input image, while encountering fixed-position samples $I \in \mathcal{D}_o$, we set the position switch embedding to $E_o$.

\begin{figure*}
    \centering
    \includegraphics[width=0.98\linewidth]{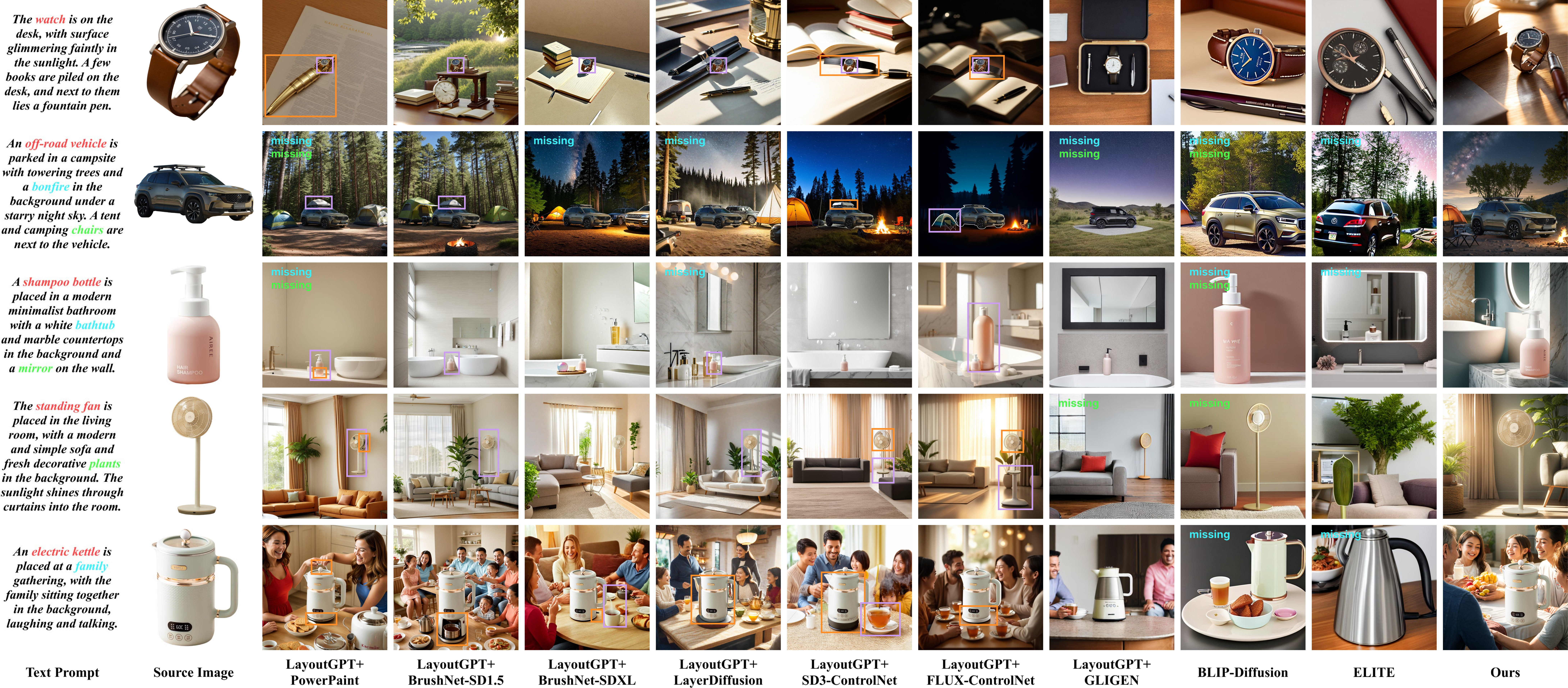}
    \vspace{-0.2cm}
    \caption{Comparison between our A$^\text{T}$A and the baseline methods. We highlight the unreasonable extension parts with \textcolor{orange}{orange} boxes and the unreasonable layouts with \textcolor{violet}{purple} boxes, and label the missing objects with \textcolor{cyan}{corresponding} \textcolor{green}{colors}. Please zoom in for more details.}
    \label{fig:comparison}
\vspace{-0.4cm}
\end{figure*}

\begin{table}[t]
\scriptsize
\centering
\setlength\tabcolsep{4pt}
\renewcommand{\arraystretch}{1.1}
\caption{Quantitative Comparisons for the Text-Guided Subject-Position Variable Background Inpainting task.}
\vspace{-0.1cm}
\label{tab:comparison}
\resizebox{\columnwidth}{!}{
\begin{tabular}{l|c|c|c|c|c}
\toprule
\multirow{2}*{Methods} & {Image Quality} & Extension Ratio & 
\yrn{Text-alignment} & \yrn{Multi-subject} & \CR{Rationality}\\
& {FID $\downarrow$} & BiRefNet $\downarrow$ & VQA Score $\uparrow$ & FlorenceV2 $\downarrow$ & \CR{GPT4o $\uparrow$} \\
\midrule
LayerDiffusion & 100.29 & 21.09\% & 0.862 & 26.40\% & 85.6\\
PowerPaint & 89.31 & 21.16\% & 0.805 & 25.04\% & 91.6\\
BrushNet-SDXL & 90.72 & 14.27\% & 0.869 & 20.32\% & 92.1\\
BrushNet-SD1.5 & 90.79 & 16.26\% & 0.857 & 25.40\% & 88.1\\ 
\CR{SD3-ControlNet}  & 81.14 & 24.23\% & \textbf{0.908} & 14.75\% & 89.8\\
\CR{FLUX-ControlNet} & 82.82 & 22.38\% & 0.903 & 15.09\% & 87.8\\
\midrule
GLIGEN & 213.70 & 46.82\% & 0.662 & 9.37\% & 88.9\\
ELITE & 160.14 & 58.36\% & 0.654 & 19.06\%  & 81.6\\
BLIP-Diffusion & 191.53 & 44.36\% & 0.524 & 8.03\% & 87.9\\
\midrule
\textbf{A$^\text{T}$A} & \textbf{79.66} & \textbf{7.92\%} & 0.883 & \textbf{4.73\%} & \textbf{92.8}\\
\bottomrule
\end{tabular}
}
\vspace{-0.4cm}
\end{table}


\section{Experiments}
\label{sec:experiment}

\subsection{\yr{Experimental} Settings}

\noindent
\textbf{Datasets.} \szm{The datasets~\cite{kuznetsova2020open,lin2014microsoft,wang2023imagen} typically employed for inpainting tasks, 
have limitations} due to their low-quality, cluttered real-world \hut{images} or \szm{imprecise} annotated mask boundaries~\cite{ju2024brushnet}, \szm{while} our task \szm{\yrn{requires training} datasets with foreground subjects that can be completely isolated, allowing for adaptive movement of the subject.}
Thus, \szm{we follow Pinco~\cite{pinco} and} collect $79$\hut{K} high-quality images with varying aspect ratios across diverse subject categories, including products, vehicles, people, and animals.
We use BiRefNet~\cite{zheng2024bilateral} to generate subject segmentation masks, and ZoeDepth~\cite{bhat2023zoedepth} to produce depth maps.
\szm{\yrn{When constructing the training samples for variable-position inpainting,} to deal with samples with different subject positions, we 
\yrn{process the images to}
move \yrn{the subject} to the center of the image, and enlarge them isometrically to $95\%$ of the maximum scale (Fig.~\ref{fig:adaptive_sample}).
The mask $m$ and depth map $d_S$ are scaled up with the same proportions.
This process results in data pairs where the input subject \yrn{(centered image)} and the reference subject \yrn{(original image)} differ in position and scale.}
We use Hunyuan-Vision~\cite{Hunyuan-Large} to label the subject category and generate detailed text prompts.

\noindent
\textbf{Implementation Details.} Our training is conducted on $16$ NVIDIA A100 GPUs for $100$ epochs, with a batch size of $4$ per GPU. 
We use AdamW~\cite{loshchilov2017decoupled} optimizer, and \szm{the learning rate is set to 
\CR{$1\text{e-}4$,}
using a $300$-steps warmup.}
We choose DiT~\cite{peebles2023scalable} as the \yrn{architecture} for the PosAgent block.
The image aspect ratios for training include $1\!:\!1$, $9\!:\!16$, $16\!:\!9$.
\CR{To ensure the full consistency of the unmasked region, previous works~\cite{zhuang2023powerpaint, ju2024brushnet} usually adopt a copy-and-paste operation.
For A$^\text{T}$A, since the subject position and scale in generated images are adaptively variable, we first use Florence-2~\cite{xiao2024florence} to detect the accurate bounding box of the subject based on its text prompt.
Then we rescale the subject image to fit the bounding box and conduct the blending operation~\cite{ju2024brushnet}.}

\noindent
\textbf{Evaluation Metrics.}
We \hut{evaluate the model performance by} the Image Quality, Extention Ratio, Text-alignment, \yrn{Multi-subject Rate}\CR{, and Position Rationality}:
1) We use FID to assess the image quality;
2) To measure the subject consistency, we utilize BiRefNet~\cite{zheng2024bilateral} to generate an accurate subject mask and calculate the OER~\cite{chen2024virtualmodel} to measure the subject extension ratio;
3) For text alignment, we use VQA Score~\cite{lin2025evaluating} to assess the subject-text alignment, and use FlorenceV2~\cite{xiao2024florence} to detect the multi-subject rate based on the given subject prompt.
\CR{Following Pinco~\cite{pinco}, We leverage GPT4o to evaluate the rationality of the subject position (full score 100).}
\yrn{\CR{4)} We further conduct a user study to assess the rationality of the subject position and overall quality.}

\subsection{Comparisons}
\szm{We \yrn{construct} baselines for} the Text-Guided Subject-Position Variable Background Inpainting task \szm{by \yrn{coupling} existing} text-guided inpainting methods
\yrn{with a layout generation method}.
\szm{
Existing \yrn{T2I inpainting} methods only achieve inpainting \yrn{with subject-position fixed}, 
but our task requires the subject to adaptively change its position.
To enable a fair comparison, we use LayoutGPT~\cite{feng2024layoutgpt} to \yrn{generate} a \yrn{plausible} \szm{subject-position layout}\yrn{, and then}}
combine LayoutGPT with those existing \yrn{inpainting methods for comparison,} 
including PowerPaint~\cite{zhuang2023powerpaint}, BrushNet1.5, BrushNetXL~\cite{ju2024brushnet}, LayerDiffusion~\cite{zhang2024transparent}\CR{, SD3-ControlNet~\cite{esser2024scalingrectifiedflowtransformers} and FLUX-ControlNet~\cite{flux2024}}. 
Specifically, we first use LayoutGPT to generate the bounding box of the subject based on the given text, then scale and shift the subject into the given region\footnote{\CR{We also compare with these methods using user-provided subject layout instead of LayoutGPT, shown in \#Appendix Sec.~D,~E.}}. We also combine LayoutGPT with GLIGEN~\cite{li2023gligen}, which can generate an image based on the given bounding box using text and conditional image. Finally, we compare \yrn{with} $2$ personalized image generation methods, ELITE~\cite{wei2023elite} and BLIP-Diffusion~\cite{li2024blip}. 

\noindent \textbf{Qualitative Comparisons.}
Fig.~\ref{fig:comparison} shows the results of qualitative comparisons with different methods\footnote{More images of different aspect ratios are shown in \yrn{\#Appendix} Sec.G.}.
\szm{As shown in Fig.~\ref{fig:comparison}}, LayoutGPT \szm{tends to} produce unreasonable layouts \szm{(the 1$^{st}$ \yrn{and 4$^{th}$} rows)} or \szm{layouts with inappropriate ratio} \yrn{(the 3$^{rd}$ row)}. In the 2$^{nd}$ row, although a suitable layout is given, these inpainting models cannot understand the position information of the subject, and the generated background results conflict with the text, resulting in inharmonious image content layout or lacking objects in the text.
\szm{Meanwhile, customized methods struggle with subject consistency and text-image alignment due to inadequate control over image features, leading to poor performance.}

\noindent \textbf{Quantitative Comparisons.}
Tab.~\ref{tab:comparison} \CR{shows} the quantitative comparisons \hut{among} different methods, \szm{where}
our A$^\text{T}$A achieves the best FID score, demonstrating high generation quality. Additionally, our extension ratio \CR{and multi-subject rate are} less than \CR{60\%} of other methods, indicating A$^\text{T}$A's ability to prevent subject expansion \CR{and redundancy}. 
\CR{Finally, 
we achieve the best position rationality score, indicating a better positional relationship.}
Overall, these metrics demonstrate our superiority in terms of generation quality, extension rate, \CR{and subject position rationality}.

\begin{figure}[t]
    \centering
    \includegraphics[width=0.95\linewidth]{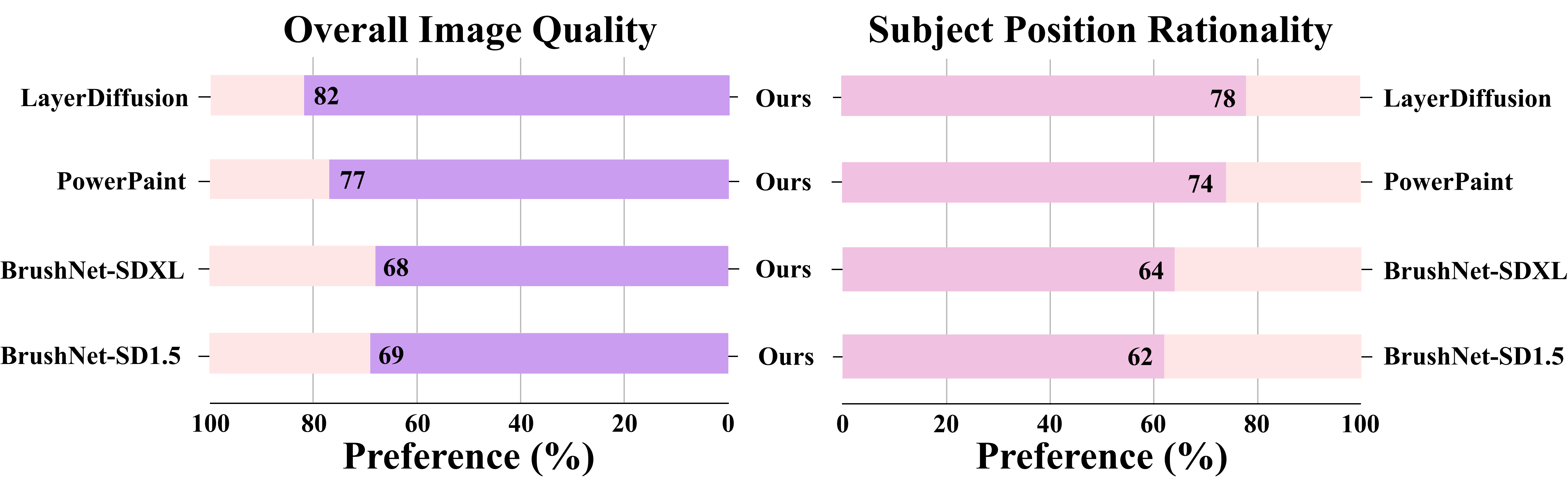}
    \vspace{-0.1cm}
    \caption{User study between A$^\text{T}$A and other inpainting methods.}
    \label{fig:user-study}
    \vspace{-0.5cm}
\end{figure}

\noindent \textbf{User Study.}
We invite $31$ participants for our user study, each receiving $40$ sets of test questions\szm{, \yrn{with} each set consisting of $2$ images: one generated by A$^\text{T}$A and the other from a different method.}
Participants are asked to choose the better image based on 
\yrn{\textit{rationality of subject position},}
and overall image quality.
As shown in Fig.~\ref{fig:user-study}, our A$^\text{T}$A receives the most preference from the participants\CR{, demonstrating our superior position rationality}.

\subsection{Ablation Study}
We conduct extensive ablation studies to \yrn{validate} the effectiveness of \yrn{each module}. \yrn{We compare with} \szm{different Displacement Transform structures and} \yrn{analyze} condition embedding injection \szm{in the PosAgent block.}

\begin{table}[t]
\scriptsize
\centering
\setlength\tabcolsep{2pt}
\renewcommand{\arraystretch}{1.1}
\caption{Ablation study on different structures for Displacement Transform (DT) module. Here, HT refers to Hadamard Transform.}
\vspace{-0.2cm}
\label{tab:dt}
\resizebox{\columnwidth}{!}{%
\begin{tabular}{l|c|c|c|c}
\toprule
\multirow{2}*{DT Module} & {Image Quality} & Expand Ratio & {Text-alignment} & \yrn{Multi-subject}\\
 & {FID $\downarrow$} & BiRefNet $\downarrow$ & VQA Score $\uparrow$ & FlorenceV2 $\downarrow$ \\
\midrule
w/o DT & 81.74 &10.89\% & 0.873 & 4.89\% \\
Separate DT & 83.83 &10.25\% & 0.880 & 4.84\% \\
Sequential DT & 83.36 &9.73\% & 0.875 & 5.34\% \\
Reverse HT & 85.88 &11.26\% & 0.877 & 5.79\% \\
Reverse DT (A$^\text{T}$A) & \textbf{79.66} &\textbf{7.92\%} & \textbf{0.883} & \textbf{4.73\%} \\ \bottomrule
\end{tabular}
}
\vspace{-0.3cm}
\end{table}

\begin{figure}[t]
    \centering
    \includegraphics[width=\linewidth]{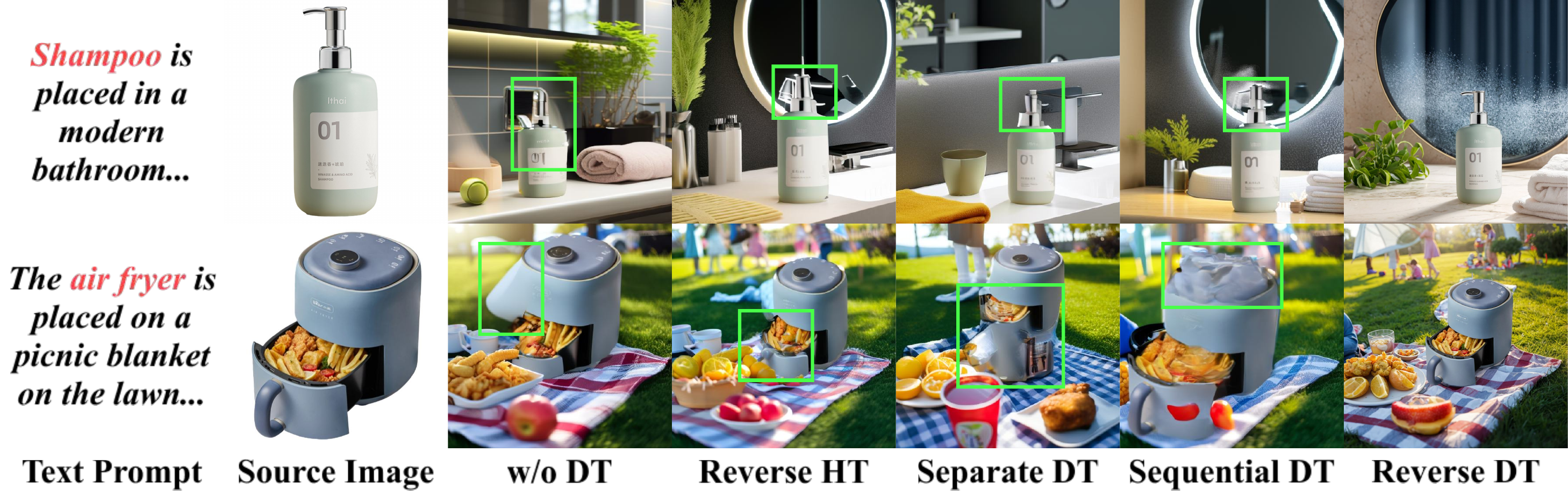}
    \caption{Qualitative ablation study on different structures for Displacement Transform (DT) module. We highlight the deformation part of the subject with \textcolor{green}{green} boxes.}
    \label{fig:dt}
    \vspace{-0.4cm}
\end{figure}

\noindent \textbf{Displacement Transform modules.}
We perform ablation studies \szm{by \yrn{comparing} various Displacement Transform (DT) structures to assess their impact (Tab.~\ref{tab:dt} and Fig.~\ref{fig:dt}).
Initially, we \yrn{remove} the \yrn{entire} DT module, relying solely on the Swin-Transformer to process condition data, 
\yrn{but the generated results have less plausible subject positions and layouts.}
Next, we examine the Separate DT module, comprising $N$ distinct PosAgent blocks, each tasked with altering a specific attention map $C_I^i$. This approach results in disjointed feature transformations across levels, causing subject distortions.
We also assess the Sequential DT module, featuring consecutive PosAgent blocks with the output of the $i$-th block $\theta_i$ affecting the $i$-th attention map $C_I^i$. This module also induces subject deformation and performs inferior to our Reverse DT.
Additionally, we replace the SFT with the Hadamard transform and modify the hierarchical features using the Hadamard product, resulting in noticeable subject deformation. All these findings underscore the superiority of our Reverse Displacement Transform in maintaining subject integrity and overall image quality.}

\begin{figure}[t]
    \centering
    \includegraphics[width=\linewidth]{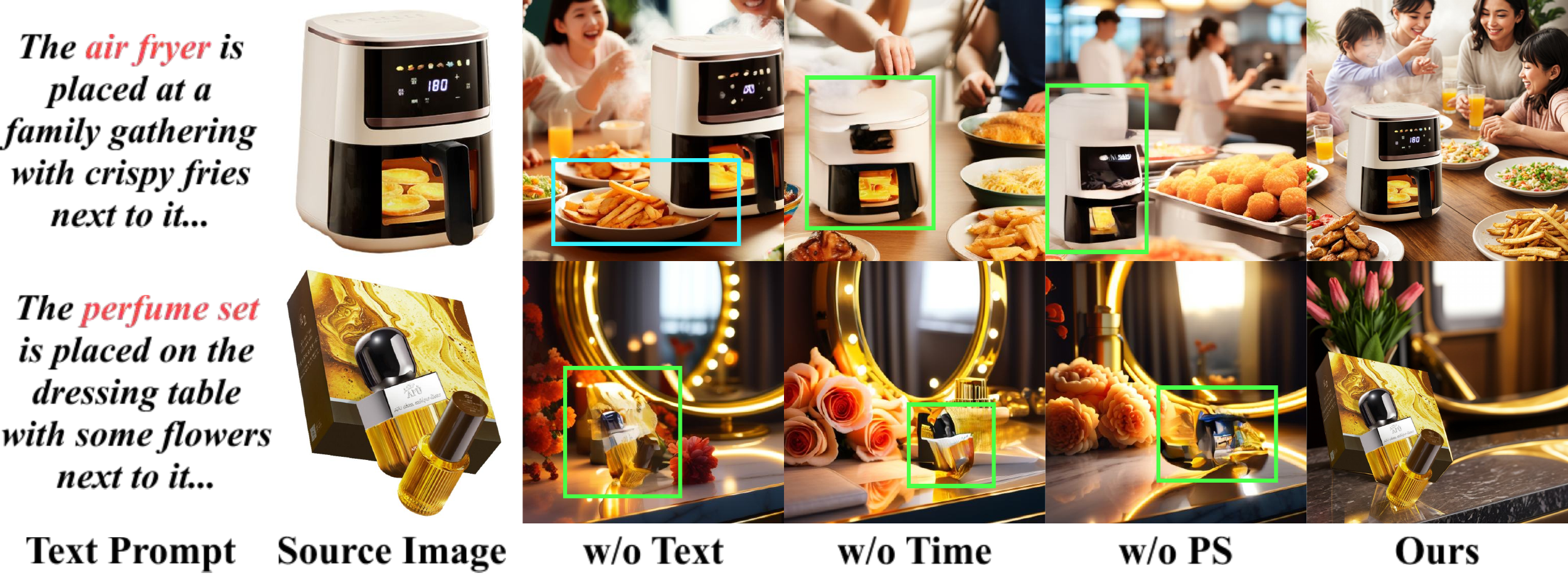}
    \caption{Qualitative ablation study on the conditions of the PosAgent module. We highlight the deformation part of the subject with \textcolor{green}{green} boxes and the misplacement part with the \textcolor{cyan}{blue} box.}
    \label{fig:pos}
    \vspace{-0.4cm}
\end{figure}

\noindent \textbf{Condition Injected to PosAgent blocks.}
\szm{We perform ablation studies on \yrn{the conditions input to} the PosAgent block, \yrn{\emph{i.e.,}} time, text, and position switch embeddings}, 
\yrn{to verify their roles,}
\szm{as illustrated in Tab.~\ref{tab:pos} and Fig.~\ref{fig:pos}.}
1) \szm{Given that the Position Switch is integral} to our hybrid training strategy, we train the PosAgent block w/o Position Switch input using the subject-position variable images $\mathcal{D}_v$ \szm{only, and we unexpectedly observed 
\yrn{unstable performance and worse quantitative metrics than w/ the switch}}.
\szm{We attribute this to the hybrid training strategy's ability to effectively separate feature extraction and \yrn{displacement} transform processes, 
\yrn{and enhance the ability in each task (position-variable/fixed) by multi-task learning.}}
2) \szm{When the PosAgent lacks time embedding,} the resulting images \szm{occasionally exhibit deformation,} and the metrics indicate a decline in performance. \szm{Moreover, omitting time embedding in PosAgent leads to an unstable training process.}
3)
\szm{In the case of the PosAgent without text embedding}, \yrn{the image quality, extension ratio, and text-alignment are worse, whereas}
we \szm{notice a slight improvement in the multi-subject rate.}
The potential reason is that\szm{, without text embedding,} the transformed feature\szm{s retain more subject information at the image level,} allowing \szm{the base model} to recognize the injected data \szm{more effectively}.
However, \szm{the absence of text guidance in PosAgent makes it challenging to position subjects appropriately in line with textual prompts, potentially leading to missing objects or misplacement due to conflicts between subject positioning and the text.}

\begin{table}[t]
\scriptsize
\centering
\setlength\tabcolsep{2pt}
\renewcommand{\arraystretch}{1.1}
\caption{Ablation study on the conditions of the PosAgent module. Here, PS refers to Position Switch.}
\vspace{-0.2cm}
\label{tab:pos}
\resizebox{\columnwidth}{!}{%
\begin{tabular}{ccc|c|c|c|c}
\toprule
\multicolumn{3}{c|}{PosAgent} & {Image Quality} & Expand Ratio & {Text-alignment} & \yrn{Multi-subject}\\
Text & Time & PS & {FID $\downarrow$}  & BiRefNet $\downarrow$ & VQA Score $\uparrow$ & FlorenceV2 $\downarrow$ \\
\midrule
\checkmark & \checkmark &  & 83.71 & 9.68\% & 0.879 & 5.95\% \\
\checkmark &  & \checkmark & 82.20 & 8.41\% & 0.879 & 4.89\% \\
 & \checkmark & \checkmark & 82.36 & 8.08\% & 0.881 & \textbf{4.61\%} \\
\checkmark & \checkmark & \checkmark & \textbf{79.66} & \textbf{7.92\%} & \textbf{0.883} & \underline{4.73\%} \\ \bottomrule
\end{tabular}
}
\vspace{-0.3cm}
\end{table}

\section{Conclusion}
\label{sec:conclusion}
This paper introduces a novel Text-Guided Subject-Position Variable Background Inpainting task, aiming to generate inpainted images with a harmonious positional relationship between the subject and the background.
For this task, we propose the Adaptive Transformation Agent (A$^\text{T}$A) framework, which dynamically adjusts the subject's position relative to the background.
A$^\text{T}$A leverages multiple PosAgent blocks for displacement prediction, the RDT module for feature transform, and a Position Switch Embedding for flexible positioning control.
Experiments confirm the effectiveness of our A$^\text{T}$A in both variable and fixed subject-position inpainting tasks.

\section*{Acknowledgments}
This work was supported by Shanghai Sailing Program (22YF1420300), National Natural Science Foundation of China (No. 62302297, 72192821, 62272447, 62472282, 62472285), Young Elite Scientists Sponsorship Program by CAST (2022QNRC001), the Fundamental Research Funds for the Central Universities (YG2023QNB17, YG2024QNA44), Beijing Natural Science Foundation (L222117).
{
    \small
    \bibliographystyle{ieeenat_fullname}
    \bibliography{main}
}
\clearpage
\appendix
\section*{\LARGE Appendix}


\section{Overview}

In this supplementary material, we mainly present the following \yrn{contents}:
\begin{itemize}
    \item More technical details of our A$^\text{T}$A  model structure \yrn{(Sec.~\ref{sec:model_structure})};
    \item More technical details of the evaluation metrics \yrn{(Sec.~\ref{sec:evaluation_metrics})};
    \item More qualitative comparison for the subject-position variable inpainting task \yrn{(Sec.~\ref{sec:comparison_pos_variable})};
    \item Qualitative comparison for the subject-position fixed inpainting task 
    \CR{with user-provided layout}
    \yrn{(Sec.~\ref{sec:comparison_pos_fixed})};
    \item More \yrn{details of} the user study \yrn{(Sec.~\ref{sec:user_study})};
    \item More results of A$^\text{T}$A \yrn{with} different aspect ratios \yrn{(Sec.~\ref{sec:results_aspect_ratios})};
    \item Visualization and analysis of the attention map \yrn{(Sec.~\ref{sec:visualization_attention_map})};
    \item \CR{More analysis of the RDT module (Sec.~\ref{sec:analysis_rdt});}
    \item \CR{Prospect of future works (Sec.~\ref{sec:prospect});}
    \item Image copyright (Sec.~\ref{asec: cpr}).

\end{itemize}

\section{\yrn{More Details of} Model Structure}
\label{sec:model_structure}
In this section, we provide \yrn{a} more detailed description of the model structure. We \yrn{adopt} the architecture Hunyuan-DiT-g/2~\cite{li2024hunyuan} as our \yrn{base model}, which has $40$ blocks and an embedding dimension of $1,408$.
Our A$^\text{T}$A model consists of $4$ modules: Feature extraction, Reverse displacement transform, Feature fusion, and Diffusion denoising \yrn{(refer to the main paper Fig. 2 for overall architecture)}.

\noindent
\textbf{Feature Extraction.} We use a tiny Swin-Transformer~\cite{liu2021swin} backbone as \yrn{the} feature extractor \yrn{for the subject image}, and pre-process the input with an $8\times 8$ window size. This process yields a set of multi-scale feature maps, denoted as $\{C^1_I, \cdots, C^N_I\}$, with $N=4$ and the dimensions of $C^1_I$ being $C \times H \times W$. Following each Swin-Transformer~\cite{liu2021swin} stage, the number of channels is doubled compared to the previous stage, while the spatial dimensions (height and width) are halved, resulting in a new \yrn{feature map} of $2C \times H/2 \times W/2$ \yrn{dimension}. 
\yrn{Since the $4$ subject feature maps will be injected into the subject cross-attention module (after the displacement transform), which requires the injected feature to have a unified dimension, we use}
a convolutional layer to adjust the features, 
\yrn{mapping them to the same dimension for future injection.}

\begin{figure}[t]
    \centering
    \includegraphics[width=\linewidth]{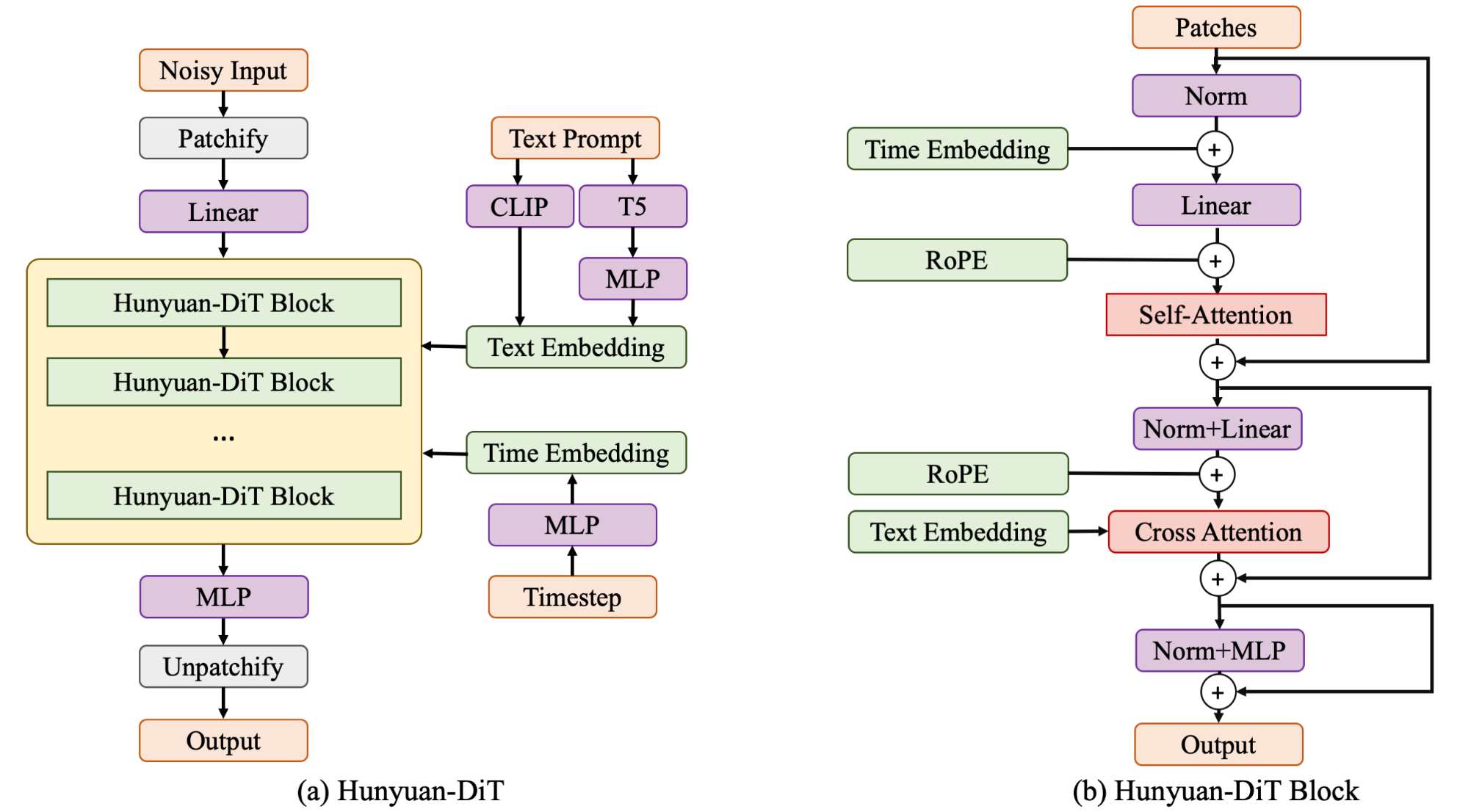}
    \caption{The model structure of Hunyuan-DiT~\cite{li2024hunyuan} \yrn{(the base model of our A$^\text{T}$A)}.}
    \label{fig:hunyuan}
    \vspace{-0.1in}
\end{figure}

\noindent
\textbf{Reverse Displacement Transform}. Given that our foundational model is based on a DiT framework, we opt for the DiT block~\cite{peebles2023scalable} enhanced with Adaptive Layer Norm (AdaLN) to serve as the architecture for the PosAgent block $\mathcal{A}_i$. The PosAgent block takes the output from the Swin-transformer $\Phi$ as its input, incorporating time \yrn{embedding} $t$, text \yrn{embedding} $C_T$, and positional switch embedding $C_P$ on the conditional side, fusing them with the input via a Layer Normalization technique. The output from each of these blocks is then utilized to perform a reverse transformation on the feature map $\{C^1_I, \cdots, C^N_I\}$ initially extracted by the Swin-transformer $\Phi$, effectively restructuring it in a reverse manner. \emph{I.e.}, the transformation parameters predicted by the first PosAgent block $\mathcal{A}_1$ are used to transform the last (deepest) feature map $C^N_I$, while the last PosAgent block $\mathcal{A}_N$ will transform the first (shallowest) feature map $C^1_I$.

\noindent
\textbf{Feature Fusion.} We employ a pre-trained cross-attention mechanism as our subject cross-attention module, which is initialized with identical weights \yrn{to} the base model\yrn{'s} cross-attention \yrn{module}. Subsequently, the output from this attention module is combined with the output from the original self-attention module using a trainable $\tanh$ weight. This setup allows the model to dynamically adjust the influence of these two attention modules depending on the specific requirements of various Hunyuan-DiT blocks.

\noindent
\textbf{Diffusion Denoising.} After all features are fused, A$^\text{T}$A performs a T-step denoising process as Hunyuan-DiT \yrn{(structure shown} in Fig.~\ref{fig:hunyuan}) and obtains the inpainted image.

\section{Evaluation Metrics}
\label{sec:evaluation_metrics}
In this section, we provide \yrn{a} more detailed description of the evaluation metrics.
We \yrn{quantitatively} evaluate the performance of the model from \CR{the following} perspectives: Image Quality, Extension Ratio, Text-alignment, Multi-subject Rate\CR{, and Position Rationality}. 

\textbf{1) Image Quality}: We calculate the Fr\'{e}chet Inception Distance (FID)~\cite{heusel2017gans} score on the MSCOCO~\cite{lin2014microsoft} dataset to evaluate the quality of generated images.

\textbf{2) Extension Ratio}: To evaluate the subject extension ratio, we adopt the OER~\cite{chen2024virtualmodel} metric which calculates the consistency between \yrn{the foreground subject} mask of the generated image against the \yrn{ground truth subject mask.} 
Specifically, we first use BiRefNet~\cite{zheng2024bilateral} to segment the generated image and obtain an accurate mask $M$ of the foreground subject.
For subject-position fixed background inpainting methods\yrn{, since the subject's position is expected to be the same position in the original image,}
\yrn{the subject mask of the original image $M_o=1-m$ serves as the ground truth subject mask. Then}
the OER score can be computed \yrn{as follows:} 
\begin{equation}
OER = \frac{\sum \text{ReLU}(M - \yrn{M_o})}{\sum \yrn{M_o}},
\end{equation}
where ReLU is the activation function.
The smaller the OER score, the better \yrn{subject consistency is achieved by} the inpainting 
model.
Since our A$^\text{T}$A can adaptively determine a suitable position and size of the subject, \yrn{and generate an inpainted image with the subject in the new position,} we utilize FlorenceV2~\cite{xiao2024florence} to detect a bounding box of the subject, and rescale the original subject mask \yrn{$M_o$} to fit the detected bounding box as the \yrn{ground truth} subject mask \yrn{$M_o'$}. 
The OER score for our A$^\text{T}$A is:
\begin{equation}
OER = \frac{\sum \text{ReLU}(M - \yrn{M_o'})}{\sum \yrn{M_o'}}.
\end{equation}

\textbf{3) Text-alignment}: To measure the text-image alignment, following Imagen3~\cite{baldridge2024imagen}, we choose VQA~\cite{lin2025evaluating} score which evaluates the alignment between an image and a text prompt by using a visual-question-answering model to answer simple yes-or-no questions about the image content.

\textbf{4) Multi-subject Rate}: In the text-guided background inpainting task, sometimes the model will repeatedly draw the given foreground subject since it cannot recognize it in the prompt. 
In this case, the generated result will contain multiple similar subjects, which may not be what the user wants.
As a result, we adopt FlorenceV2~\cite{xiao2024florence} to \yrn{detect and} count the number of the subject in the generated result given the subject name. 
Then we calculate the \yrn{ratio} that the number of the detected subject is greater than \yrn{the desired number of the subject} as \yrn{the} multi-subject rate.

\CR{\textbf{5) Postion Rationality}:
We leverage GPT-4o to evaluate each image based on placement, size, and spatial relationships. 
Each image is scored with a maximum score of 100, and the average score determines the rationality.
}

\section{\yrn{More Comparisons on} Subject-Position Variable Inpainting}
\label{sec:comparison_pos_variable}

Fig.~\ref{fig:variable-comparison} presents additional qualitative \yrn{comparisons} against various methods \yrn{on the Subject-Position Variable Inpainting task}. 
Consistent with \yrn{the results in the main paper}, \yrn{when combining previous inpainting methods with} LayoutGPT \yrn{for subject-position variable inpainting, they} often generate \yrn{results with} layouts that are either illogical or have incorrect proportions. Even when a suitable layout is produced, \yrn{previous} inpainting \yrn{methods} fail to grasp subject positioning, causing background-text conflicts or missing objects. 
\yrn{In contrast, our method A$^\text{T}$A generates high-quality inpainted results with a harmonious positional relationship between the subject and the inpainted background, as well as good text-background alignment.}

\begin{figure}[t]
    \centering
    \includegraphics[width=1.0\linewidth]{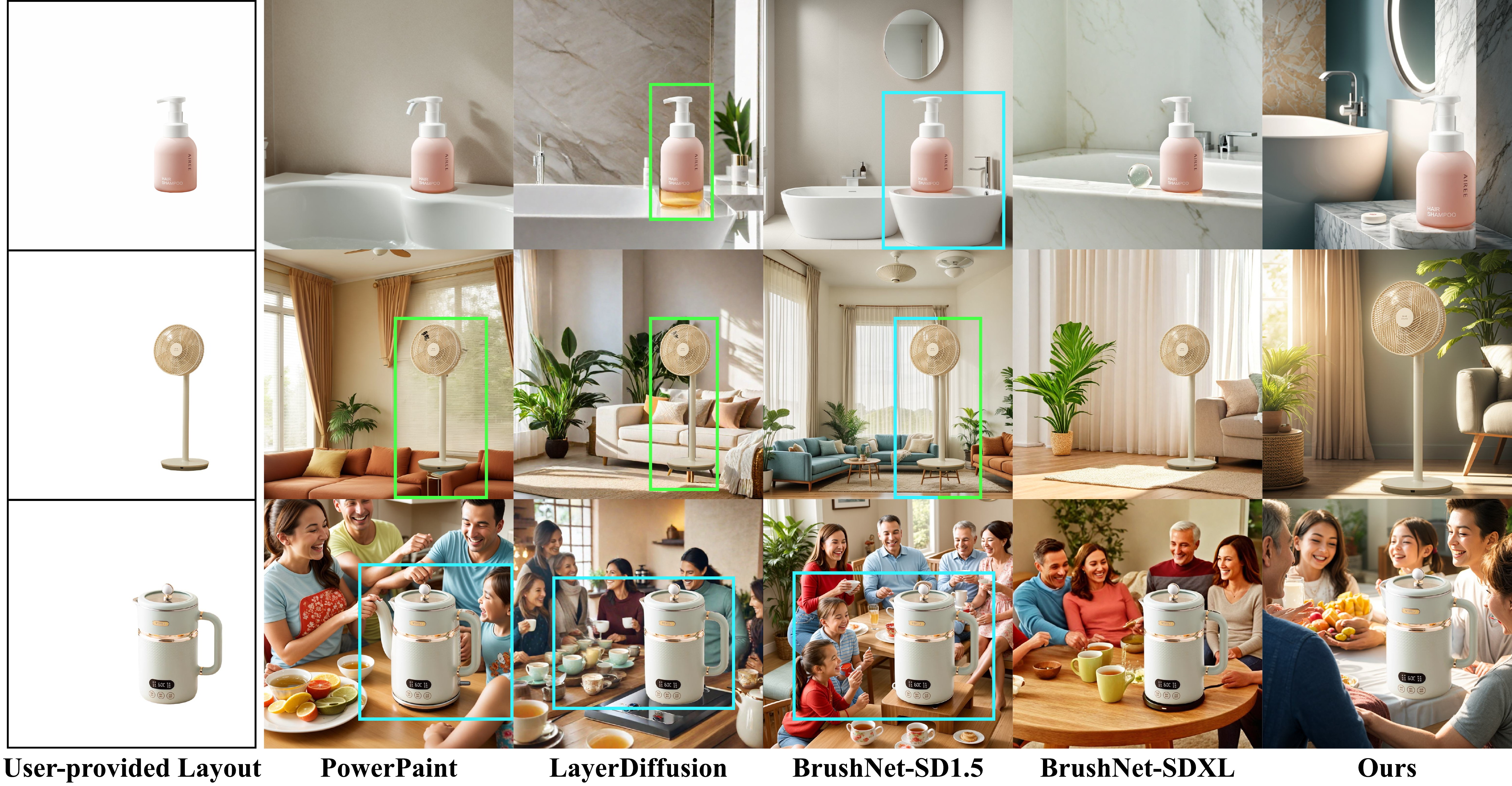}
    \caption{\CR{Comparison between our A$^\text{T}$A, and existing methods using user-provided subject layout. The comparison methods exhibit unsuitable relative \textcolor{cyan}{sizes} and \textcolor{green}{positional} relationships between foreground and background; while our method achieves harmonious positional relationships.}}
    \label{fig:user_provided_layout}
\end{figure}

\CR{Furthermore, to eliminate the influence of LayoutGPT and facilitate fairer comparison, for the methods combined with LayoutGPT in main paper Fig.~5 and Fig.~\ref{fig:variable-comparison}, 
we replace LayoutGPT with the user-provided subject layout. 
As shown in Fig.~\ref{fig:user_provided_layout}, the comparison methods 
exhibit unsuitable relative \textcolor{cyan}{sizes} and \textcolor{green}{positional} relationships between foreground and background.}

\section{\yrn{Comparisons on} Subject-Position Fixed Inpainting \CR{with User-provided Layout}}
\label{sec:comparison_pos_fixed}

\yrn{As mentioned in the main paper, our method is capable of both subject-position variable and subject-position fixed inpainting, which can be flexibly switched by setting the position switch embedding. 
To evaluate the performance on the} Subject-Position Fixed Inpainting \yrn{task,}
we have undertaken an exhaustive qualitative analysis, 
\CR{with the same user-provided subject layout for both comparison methods and our method.}
As depicted in Fig.~\ref{fig:fixed-comparison}, our approach A$^\text{T}$A demonstrates its effectiveness by \yrn{generating} satisfactory inpainting \yrn{results} when the position \yrn{of the subject} is fixed. 
\yrn{The previous inpainting methods suffer from problems including missing certain objects in the background (not well aligned with text), subject expansion, multiple subjects (of inaccurate number), and inappropriate subject size or subject mispositioned.}
\yrn{In contrast,} A$^\text{T}$A \yrn{generates high-quality results} with minimal occurrences of subjects expanding beyond the boundaries or multiple subjects\yrn{, and achieves a good text-background alignment}. Moreover, our method excels \yrn{at} creating a more \yrn{harmonious} visual relationship between the generated background and the subject.

\section{\yrn{Details of} User Study}
\label{sec:user_study}
\yrn{We conduct a user study to assess the rationality of the subject position and overall quality of the inpainted results.}
In the user study, we invited 31 
participants \yrn{majoring in computer science} to conduct the experiment, and each participant received 40 set\yr{s} of test questions.
Fig.~\ref{fig:user-study} presents some sample sets in the user study.
Each set of test questions consists of $2$ \yrn{inpainted} images: one generated by A$^\text{T}$A and another from a different method, along with the source image and the text prompt.
Each set of test images \yrn{are} shuffled to ensure that the questionnaire is blindly evaluated by the participants.
Participants are asked to choose the better image based on \textit{the rationality of the subject position} and overall image quality.
Then we calculate the average preference between A$^\text{T}$A and the other 4 compared methods, and the results are shown in Fig. 6 in the main \yrn{paper, where our A$^\text{T}$A receives the most preference from the participants\CR{, demonstrating our superior position rationality}}.

\section{More Results of Different Aspect Ratios}
\label{sec:results_aspect_ratios}

We conduct extensive comparisons for images with Different Aspect Ratios. As illustrated in \yrn{Figures}~\ref{fig:ratios1},~\ref{fig:ratios2},~\ref{fig:ratios3}, A$^\text{T}$A demonstrates its versatility by producing high-quality image outputs across a range of 
\yrn{aspect ratios.}

\begin{figure}[t]
    \centering
    \includegraphics[width=\linewidth]{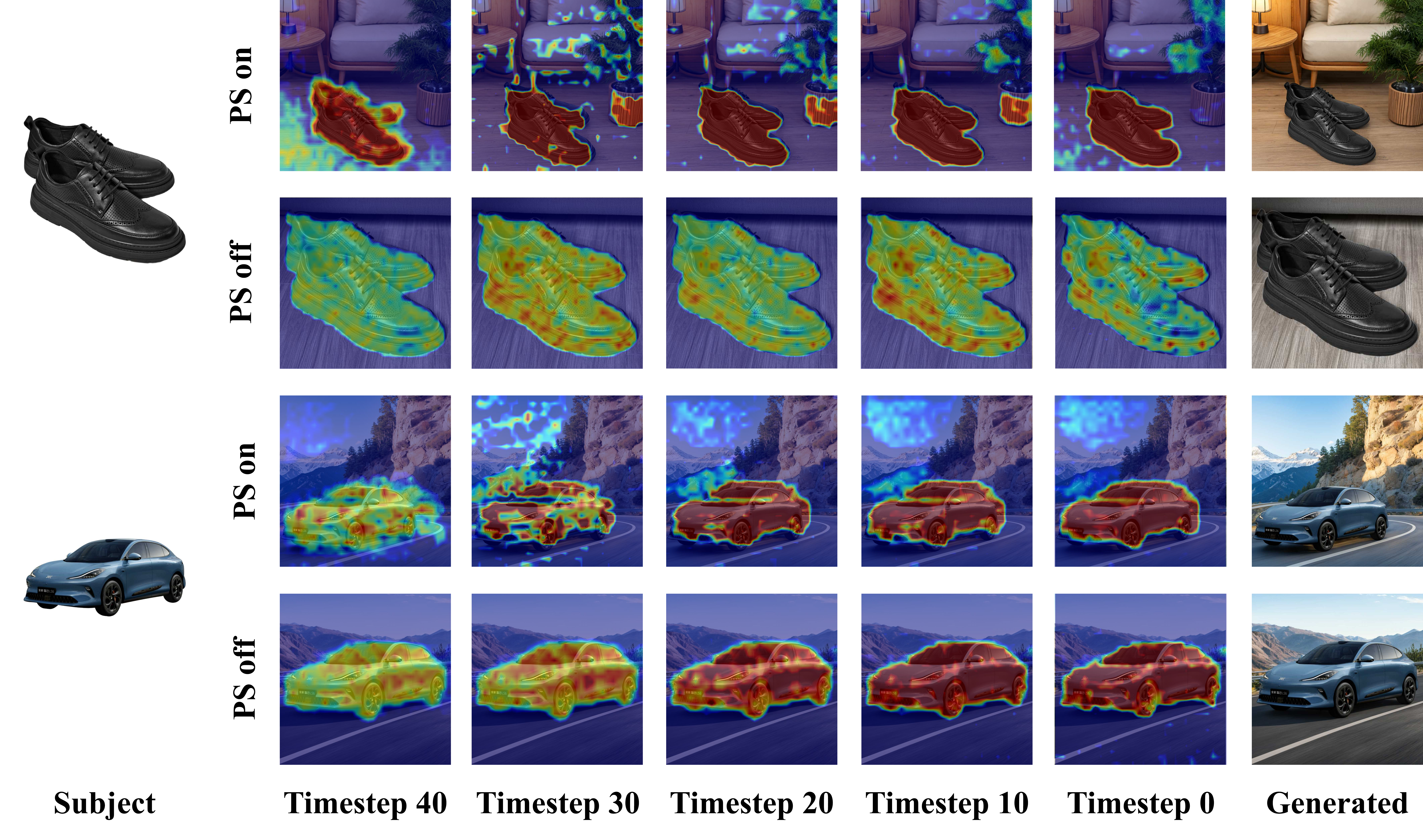}
    \caption{The \yr{visualization of} attention maps of \yr{our} model with position switch \yrn{(PS)} on and off. As the \yrn{denoising process} progresses, the model with the position switch on gradually focuses its attention \yrn{towards the} subject, \yrn{effectively adapting to the position offset;} while 
    \yrn{when the position switch is off, the model's attention remains concentrated on the subject region in the original image, without any position offset.}
    }
    \label{fig:attentionmap}
    \vspace{-0.1in}
\end{figure}

\section{Visualization \& Analysis of Attention Map}
\label{sec:visualization_attention_map}

To provide a more intuitive assessment of the position switch's performance, we perform an attention map visualization experiment. This experiment involves comparing the model's attention maps \yrn{(of the subject cross-attention layers)} with the position switch \yrn{embedding} $C_P$ activated \yrn{(subject position change enabled)} and deactivated \yrn{(subject position fixed)}. 
Fig.~\ref{fig:attentionmap} illustrates the \yrn{changes of the} attention distribution as the 
\yrn{denoising process progresses.}
When the position switch is enabled \yrn{(enabling adaptive subject position change)}, $C_P = E_v$, the model's attention gradually shifts towards the subject, effectively adapting to the position offset. 
In contrast, when the position switch is disabled \yrn{(fixing the subject position to that of the original image)}, $C_P = E_o$, the model's attention remains concentrated on the \yrn{subject region in the} original image, 
without any 
position offset. 
\yrn{Without subject-position} adaptation, \yrn{the model} can \yrn{generate} a well-inpainting image, but 
\yrn{its subject-background position relationship may not be as harmonious as}
when the position switch is \yrn{activated}.

\section{\CR{Analysis of RDT module}}
\label{sec:analysis_rdt}
In Sec. 4.3 and Sec.~\ref{sec:model_structure}, we introduce the pipeline and design details of the proposed Reverse Displacement Transform module.
In this section, we analyze the necessity of the RDT module.
The RDT module is designed to predict a suitable position for the subject, where the text information is injected through the condition channel, which is the focus of Subject-Position Variable Inpainting task.
Without RDT module, the extracted feature only contains the subject appearance information, which will lead to the generated subject being in a totally random position or even with distortions during the subjection fusion stage.
We further evaluate the additional cost brought by the RDT module, including the time cost during training stage and inference stage, as shown in Tab.~\ref{tab:performance}.
Following Pinco~\cite{pinco}, we also leverage GPT-4o to evaluate each image based on the rationality of the subject’s position (with maximum score 100). 
From Tab.~\ref{tab:performance}, we can see that the RDT module has markedly improved the position rationality score (86.2$\rightarrow$92.8) while bringing marginal time cost to the overall training (6.5\%) and inference (5.9\%) pipelines.

\begin{table}[t]
    \centering
    \scriptsize
    \setlength\tabcolsep{3pt}
    \renewcommand{\arraystretch}{1.1}
    \caption{Analysis of RDT module, where we test the time cost during training stage and inference stage and the GPT-4o score.}
    \begin{tabular}{l|c|c|c}
        \toprule
        Methods & training(s/epoch)$\downarrow$ & inference(s/image)$\downarrow$ &  rationality by GPT4o$\uparrow$ \\
        \midrule
        w/o RDT & 976.5  & 8.41 & 86.2\\
        \textbf{w/ RDT}     & 1044.5 & 8.94 & 92.8\\
        \bottomrule
    \end{tabular}
    \label{tab:performance}
\end{table}

\section{\CR{Prospect}}
\label{sec:prospect}
In our proposed ``Text-Guided Subject-Position Variable Background Inpainting task", we only focus on the position of one single subject, aiming to adaptively adjust the single subject position and generate a harmonious image.
However, for the multi-subjects scenario, it would become more complex since the relative positioning and hierarchical constraints should be taken into consideration.
Different from Subject-Position Fixed Inpainting where the multi-subjects can be combined in one image as the input, for Subject-Position Variable Inpainting, the multi-subjects should be adaptively adjusted separately, but with the constraint of relative position to align with the text prompt.
Also, the conflict of multi-subjects' positions should be avoided where the overlap of multi-subjects might happen.
In conclusion, the multi-subjects scenario is quite a meaningful but difficult direction, and our next step will consider multi-subject adaptive positioning and might adopt an additional relative position rationality module for evaluation and constraint.

\section{Copyright}\label{asec: cpr}
Some of the images presented in this paper are sourced from publicly available online resources. In our usage context, we have uniformly retained only the main subject of the original images and removed the background parts.
The copyright of all the images belongs to the original authors and brands. \textbf{The images used in this paper are solely for academic research purposes and are only used to test the effectiveness of algorithms. They are not intended for any commercial use or unauthorized distribution.}



\begin{figure*}[t]
    \centering
    \vspace{-0.1in}
    \includegraphics[width=\linewidth]{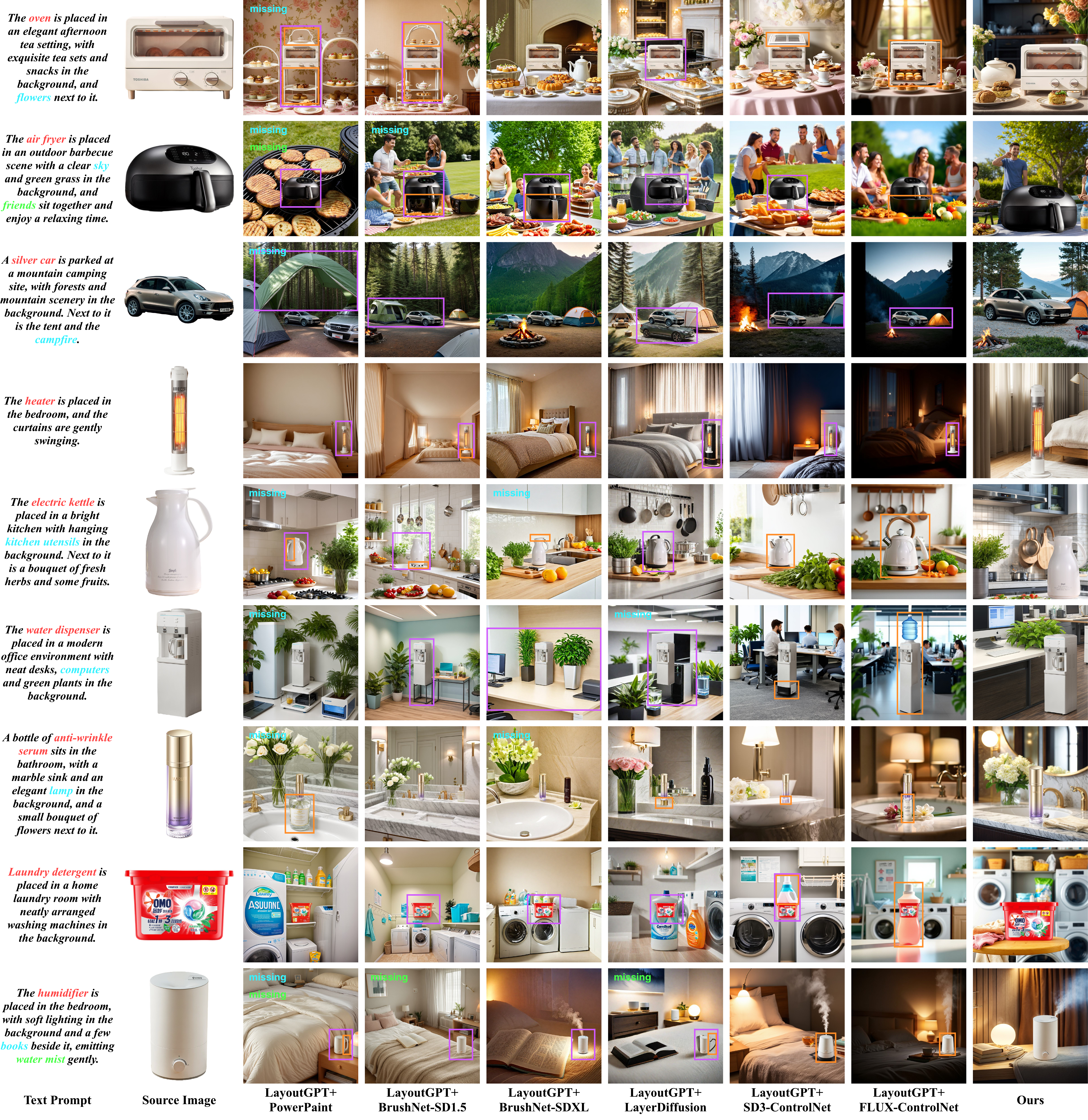}
    \vspace{-0.1in}
    \caption{\yrn{More} Qualitative results for the \textit{Subject-Position Variable} Inpainting task. \yrn{We highlight the unreasonable extension parts with \textcolor{orange}{orange} boxes and the unreasonable layouts with \textcolor{violet}{purple} boxes, and label the missing objects with \textcolor{cyan}{corresponding} \textcolor{green}{colors}. Please zoom in for more details.}}
    \label{fig:variable-comparison}
\end{figure*}

\begin{figure*}[t]
    \centering
    \includegraphics[width=\linewidth]{suppl_fig/Comparison_fixed_new.pdf}
    \caption{\yrn{More} Qualitative results for the \textit{Subject-Position Fixed} Inpainting task. \yrn{We highlight the unreasonable extension parts with \textcolor{orange}{orange} boxes, the unreasonable layouts with \textcolor{violet}{purple} boxes and the multi-subjects with \textcolor{blue}{blue} boxes. 
    Please zoom in for more details.}}
    \label{fig:fixed-comparison}
\end{figure*}

\begin{figure*}
    \centering
    \includegraphics[width=0.95\linewidth]{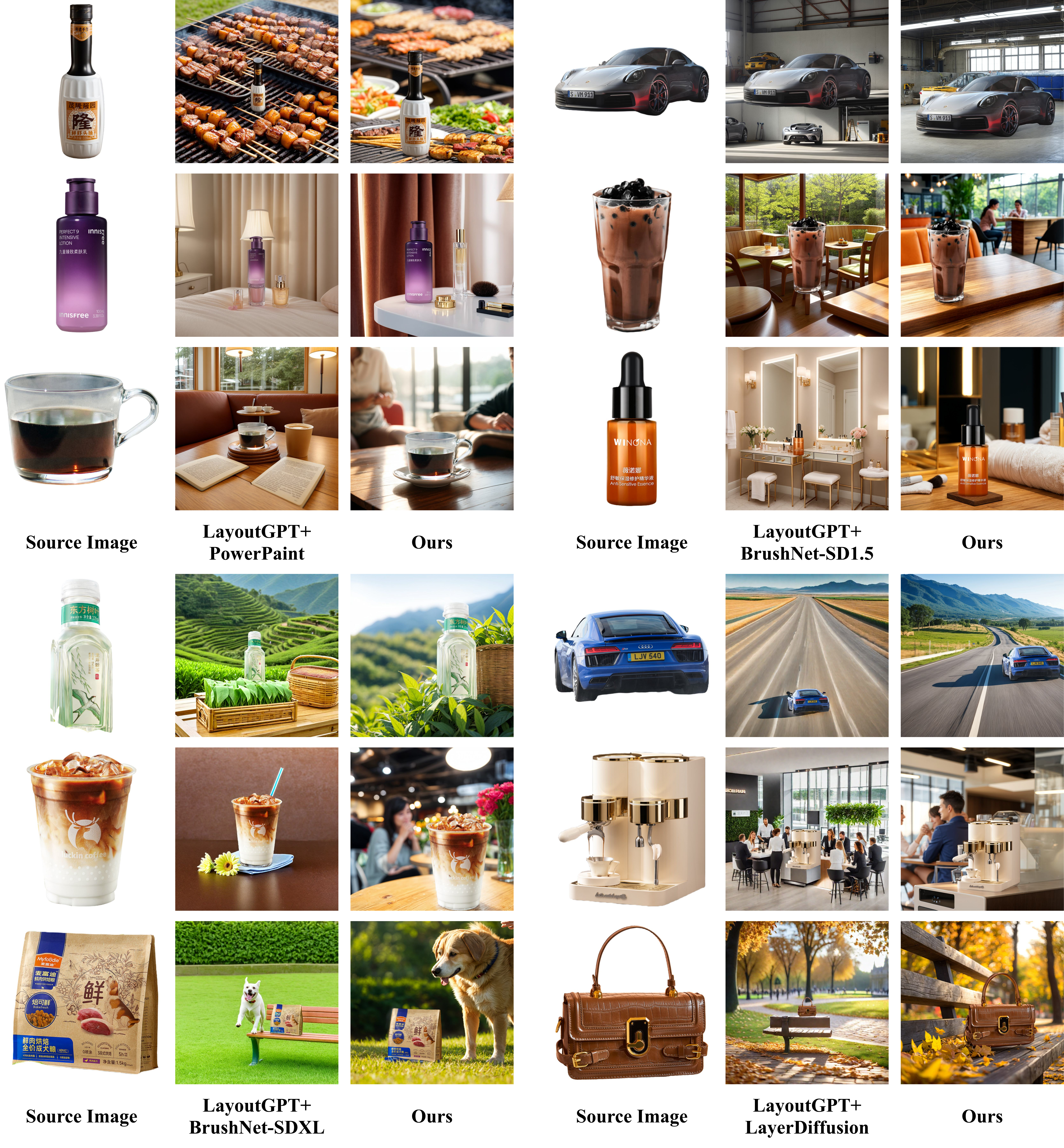}
    \caption{Some example sets used in our user study. Here we 
    \yrn{show} the names of methods \yrn{for ease of visual comparison}. However, 
    in the actual user study, the order of the images was shuffled and participants did not know the names of the methods.}
    \label{fig:user-study}
\end{figure*}

\begin{figure*}
    \centering
    \includegraphics[width=\linewidth]{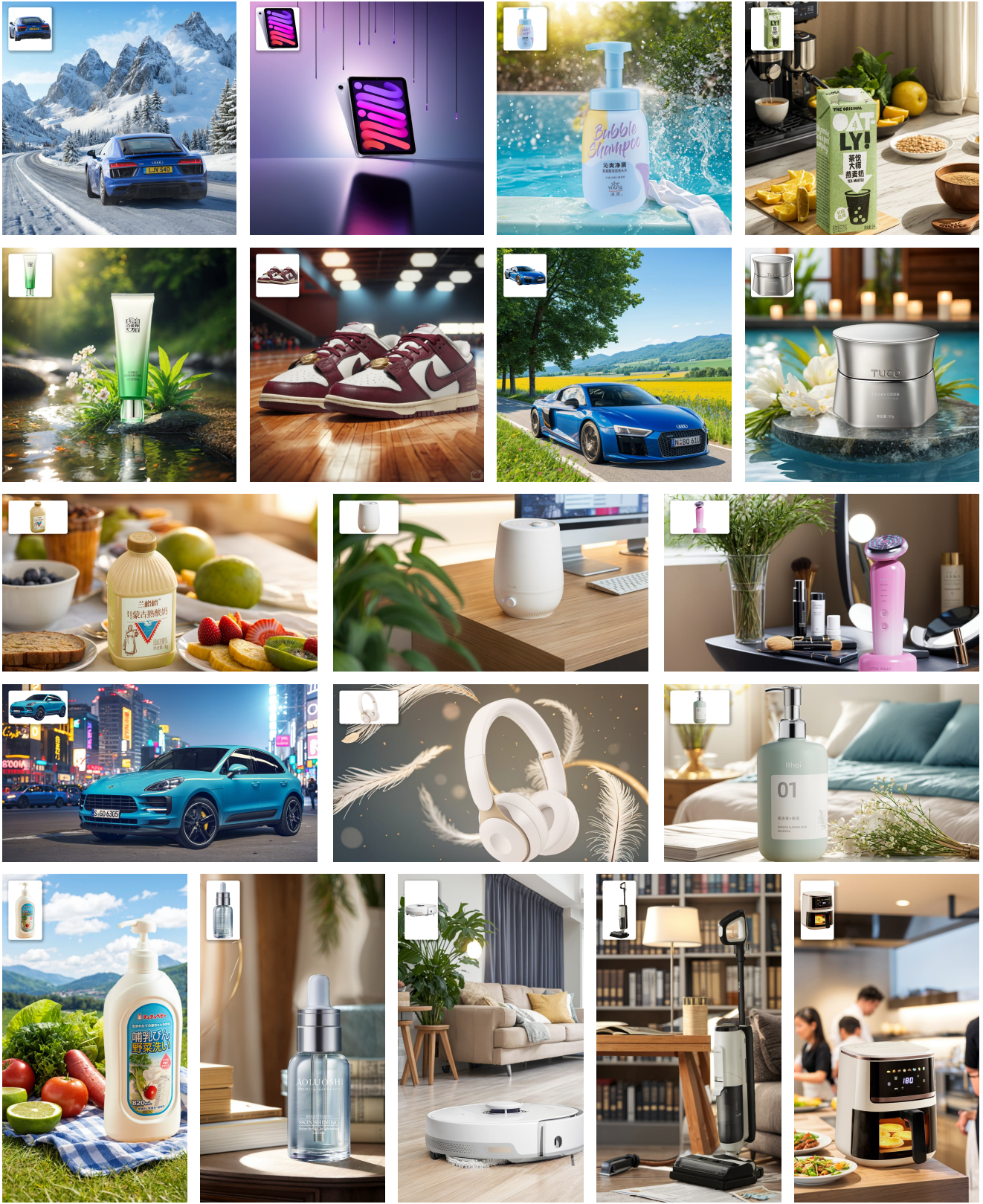}
    \caption{
    \yrn{Our A$^\text{T}$A can generate high-quality inpainted results across different aspect ratios.}}
    \label{fig:ratios1}
\end{figure*}

\begin{figure*}
    \centering
    \includegraphics[width=\linewidth]{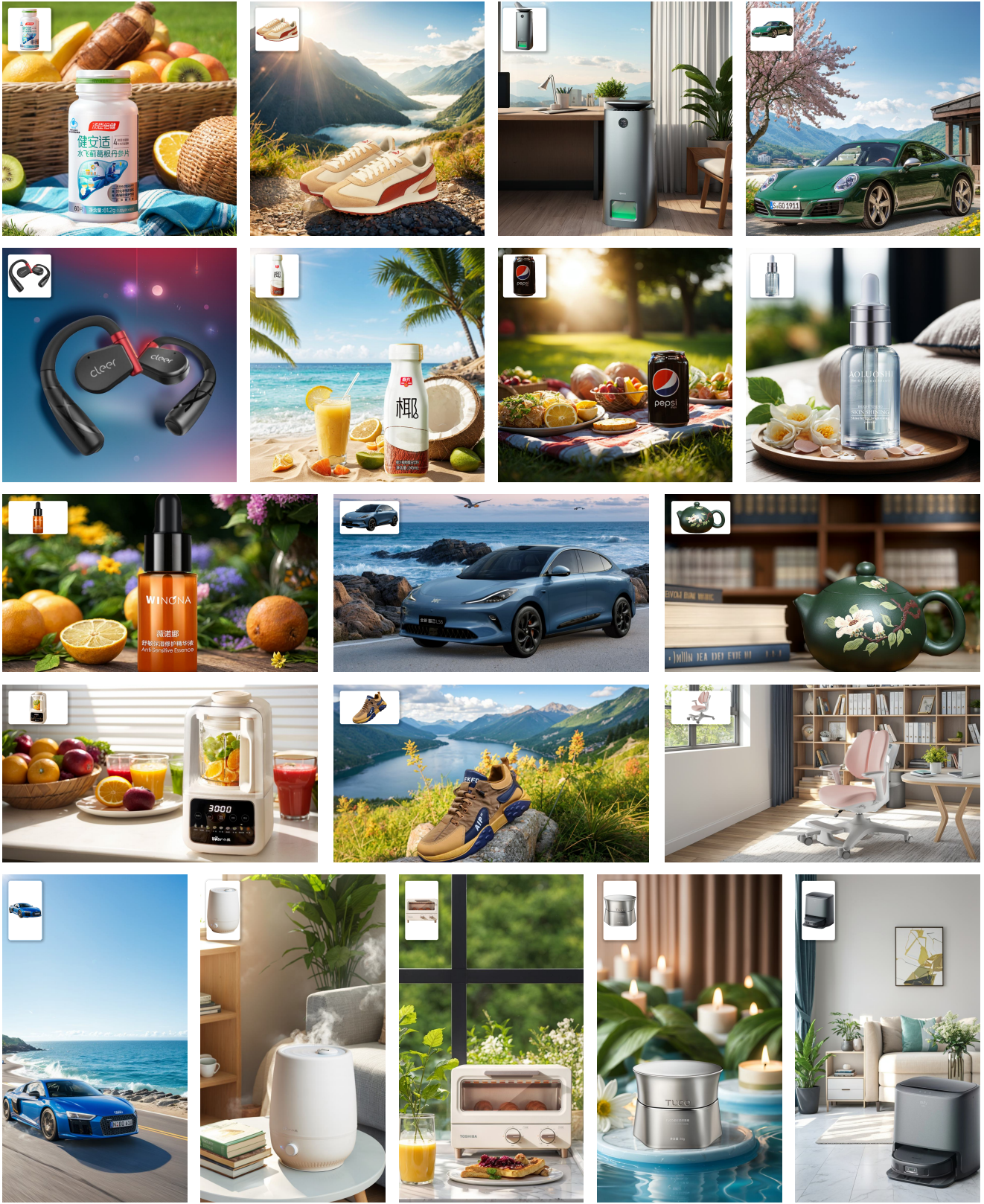}
    \caption{\yrn{Our A$^\text{T}$A can generate high-quality inpainted results across different aspect ratios.}}
    \label{fig:ratios2}
\end{figure*}

\begin{figure*}
    \centering
    \includegraphics[width=\linewidth]{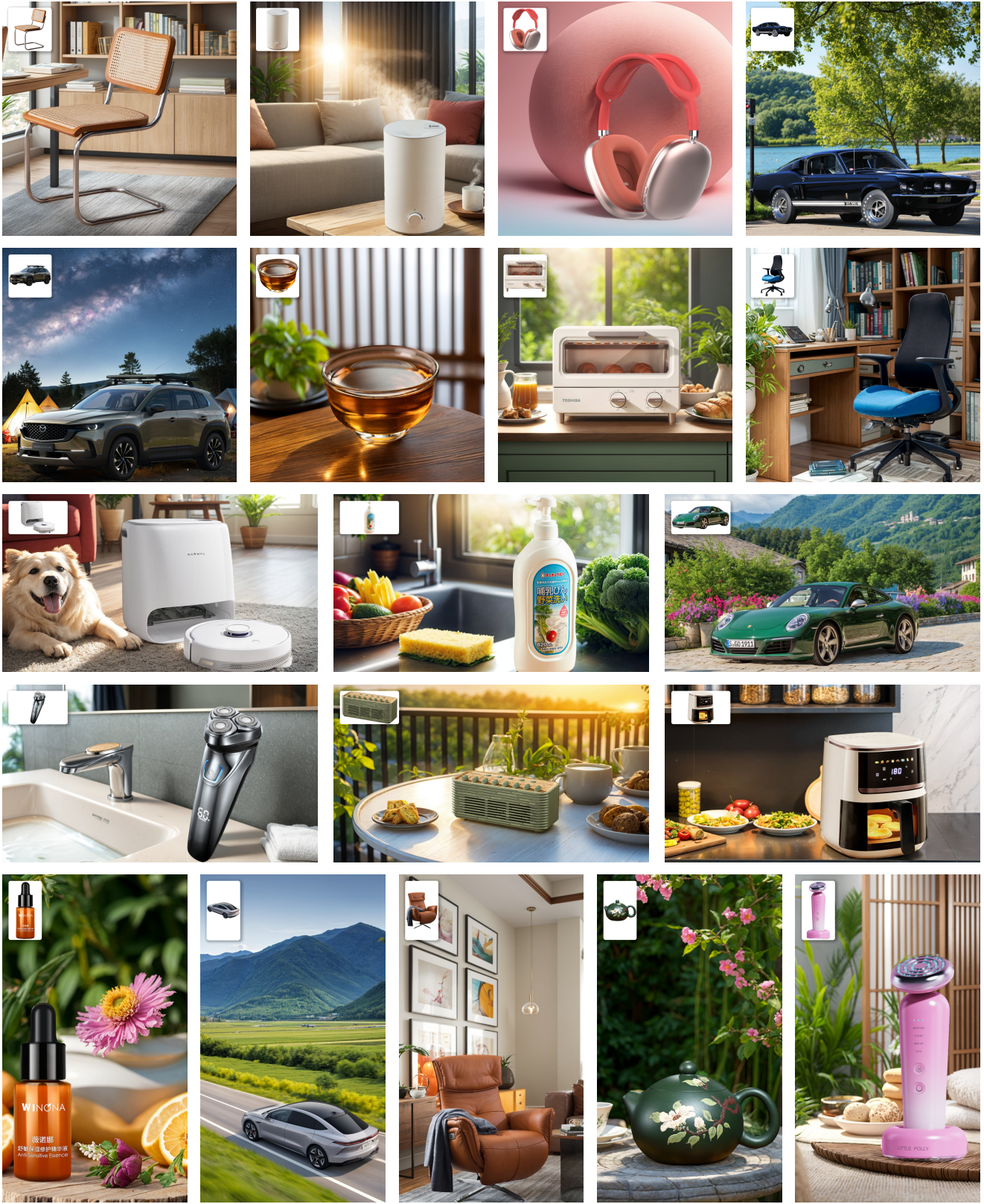}
    \caption{\yrn{Our A$^\text{T}$A can generate high-quality inpainted results across different aspect ratios.}}
    \label{fig:ratios3}
\end{figure*}


\end{document}